%% file: main.tex
\documentclass[10pt]{article} %
\usepackage[preprint]{tmlr}

\input{math_commands.tex}

\usepackage{hyperref}
\usepackage{url}
\usepackage{graphicx}
\usepackage{amsthm}
\usepackage{cleveref}
\usepackage{tikz}
\usetikzlibrary{arrows.meta,positioning,calc,shapes.geometric}
\usepackage{booktabs}
\usepackage{array}
\usepackage{amsmath, amssymb}
\usepackage[table]{xcolor}
\usepackage{tabularx}

\definecolor{gfcrG}{HTML}{D9F2FF} %
\definecolor{gfcrF}{HTML}{FFE0E0} %
\definecolor{gfcrC}{HTML}{E6E1FF} %
\definecolor{gfcrR}{HTML}{DFF6E0} %
\definecolor{gfcrHead}{HTML}{F2F2F2} %

\definecolor{gfcrHead}{RGB}{235,235,235}
\definecolor{gfcrGbg}{RGB}{235,245,255}
\definecolor{gfcrFbg}{RGB}{240,250,240}
\definecolor{gfcrCbg}{RGB}{255,245,235}
\definecolor{gfcrRbg}{RGB}{245,240,255}

\newcolumntype{Y}{>{\raggedright\arraybackslash}X}
\newcommand{\gfcrsection}[2]{%
  \rowcolor{#1}%
  \multicolumn{3}{@{}l@{}}{\textbf{#2}}\\
}

\newcommand{\GFCRG}{\cellcolor{gfcrG}\textbf{Generate}}
\newcommand{\GFCRF}{\cellcolor{gfcrF}\textbf{Filter}}
\newcommand{\GFCRC}{\cellcolor{gfcrC}\textbf{Control}}
\newcommand{\GFCRR}{\cellcolor{gfcrR}\textbf{Replay}}

\title{Generate, Filter, Control, Replay: A Comprehensive Survey of Rollout Strategies for LLM Reinforcement Learning}

\author{%
    \name Rohan Surana$^1$, 
    \name Gagan Mundada$^1$, 
    \name Xunyi Jiang$^1$, 
    \name Chuhan Wang$^1$,
    \name Zhenwei Tang$^3$, 
    \name Difan Jiao$^3$, 
    \name Zihan Huang$^1$, 
    \name Yuxin Xiong$^1$, 
    \name Junda Wu$^1$, 
    \name Sheldon Yu$^1$, 
    \name Xintong Li$^1$,
    \name Raghav Jain$^1$, 
    \name Nikki Kuang$^1$,     
    \name Sizhe Zhou$^6$ %
    \name Bowen Jin$^6$, %
    \name Zhendong Chu$^4$,     
    \name Tong Yu$^2$, 
    \name Ryan Rossi$^2$, 
    \name Kuan-Hao Huang$^5$, 
    \name Jingbo Shang$^1$, 
    \name Jiawei Han$^6$, %
    \name Julian McAuley$^1$
    \\
    \addr $^1$ University of California, San Diego\\
    \addr $^2$ Adobe Research \\
    \addr $^3$ University of Toronto \\
    \addr $^4$ University of Virginia\\
    \addr $^5$ Texas A\&M University\\
    \addr $^6$ UIUC\\
}

\def\month{MM}  %
\def\year{YYYY} %
\def\openreview{\url{https://openreview.net/forum?id=XXXX}} %

\begin{document}
\maketitle

\input{latex/0_abstract}
\input{latex/1_intro}

\input{latex/2_related}

\input{latex/3_taxonomy_new}

\input{latex/4.1_generate}

\input{latex/4.2_filter}

\input{latex/4.3_control}
\input{latex/4.4_replay}

\input{latex/5.discussion}
\input{latex/6_conclusion}

\clearpage

\input{main.bbl}
\bibliographystyle{tmlr}
\end{document}

%% file: math_commands.tex
\usepackage{amsmath,amsfonts,bm}

\def\eqref#1{equation~\ref{#1}}

\def\1{\bm{1}}

\DeclareMathAlphabet{\mathsfit}{\encodingdefault}{\sfdefault}{m}{sl}
\SetMathAlphabet{\mathsfit}{bold}{\encodingdefault}{\sfdefault}{bx}{n}

%% file: latex/0_abstract.tex
\begin{abstract}

Reinforcement learning (RL) has become a central post-training tool for improving the reasoning abilities of large language models (LLMs). In these systems, the \emph{rollout}, the trajectory sampled from a prompt to termination, including intermediate reasoning steps and optional tool or environment interactions, determines the data that the optimizer ultimately learns from, yet rollout design is often treated as an implementation detail and underreported. This survey provides an optimizer-agnostic view of rollout strategies for RL-based post-training of reasoning LLMs. We formalize rollout pipelines with unified notation and introduce Generate–Filter–Control–Replay (GFCR), a lifecycle taxonomy that decomposes rollout pipelines into four modular and composable stages: Generate proposes candidate trajectories and topologies; Filter constructs intermediate signals via verifiers, judges, or critics; Control allocates compute and makes continuation/branching/stopping decisions under budgets; and Replay retains and reuses artifacts across rollouts without weight updates, including self-evolving curricula that autonomously generate new training tasks and data. We complement GFCR with a criterion taxonomy of reliability, coverage, and cost sensitivity that characterizes the trade-offs rollout designs must navigate. Using this framework, we synthesize methods spanning RL with verifiable rewards, process supervision, judge-based gating, guided and tree/segment rollouts, adaptive compute allocation, early-exit and partial rollouts, systems-level throughput optimization, and replay/recomposition for self-improvement. We ground the framework with case studies in math, code/SQL, multimodal reasoning, tool-using agents, and agentic skill benchmarks that evaluate skill induction, reuse, and cross-task transfer. Finally, we provide a practitioner-oriented diagnostic index that maps common rollout pathologies to GFCR modules and mitigation levers, alongside open challenges for building reproducible, compute-efficient, and trustworthy rollout pipelines.

\end{abstract}

%% file: latex/1_intro.tex
\section{Introduction}
\label{sec:intro}

\begin{figure}[ht]
    \centering
    \includegraphics[width=0.8\linewidth]{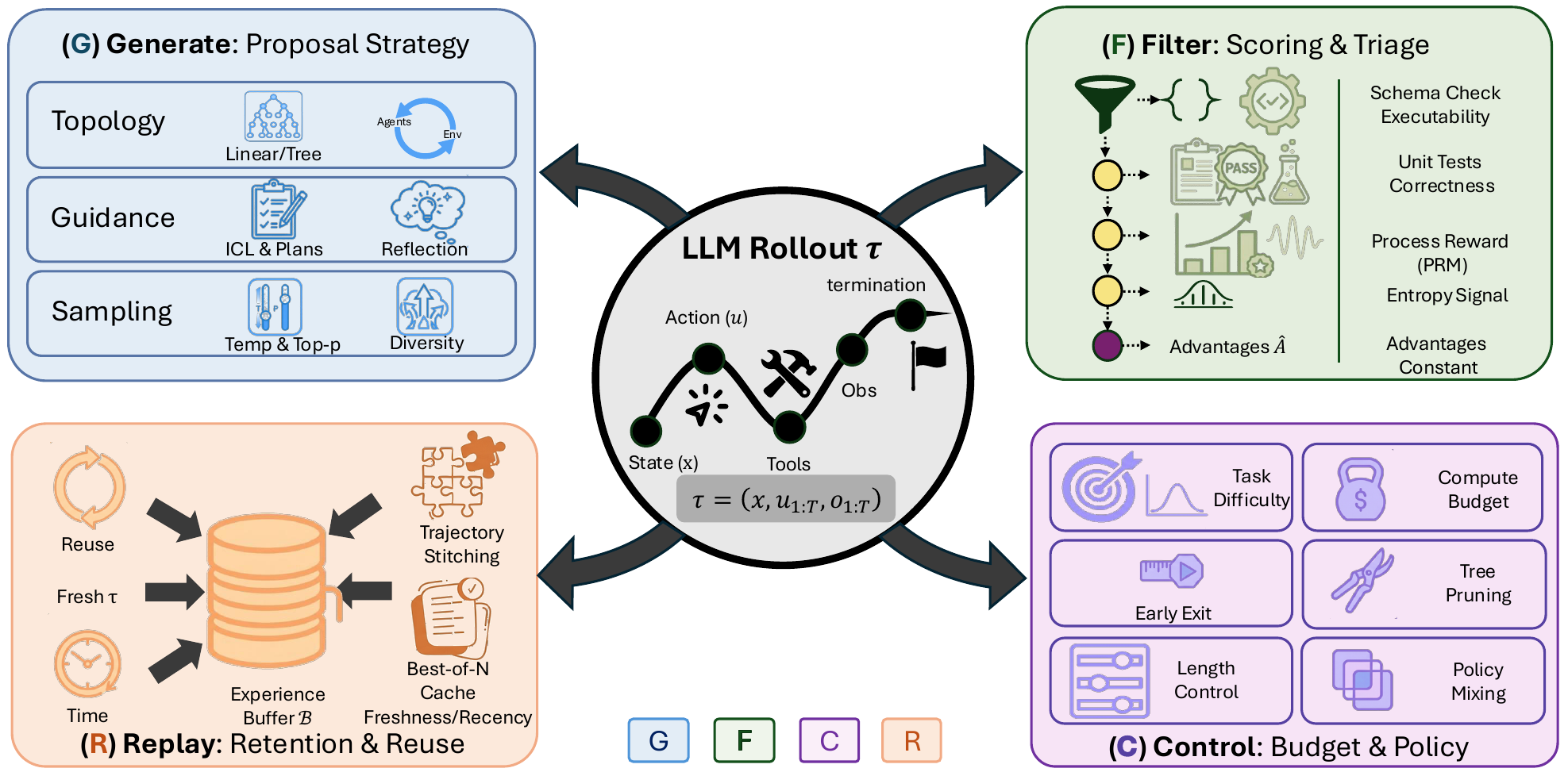}
    \caption{An overview of the rollout lifecycle and the Generate--Filter--Control--Replay (GFCR) decomposition. Rollout pipelines can be understood as modular choices about how trajectories are proposed, how intermediate signals are constructed, how compute is allocated under budgets, and what artifacts are retained and reused across rollouts.}
    \label{fig:framework-intro}
\end{figure}

Reinforcement learning (RL) has become a core component of post-training pipelines for large language models (LLMs) \citep{ouyang2022training}, enabling substantial gains on reasoning-intensive tasks such as mathematics and code/SQL generation \citep{shao2024deepseekmath,guo2025deepseek,le2022coderl,yao2026arctictext2sqlr1simplerewardsstrong} and supporting multi-step decision making in agentic settings \citep{zhang2025l0reinforcementlearninggeneral}. Much of the literature emphasizes the learning objective and optimizer (e.g., PPO/GRPO-style updates or preference-based alternatives) \citep{schulman2017proximal,shao2024deepseekmath,guo2025deepseek,rafailov2023direct,wu2025incontextrankingpreferenceoptimization} and the form of supervision (verifiers, reward models, or AI/LLM judges) \citep{christiano2017deep,ouyang2022training,lightman2023let,lee2023rlaif,zheng2023judging,gu2024llmasajudge}. However, these optimizers can only learn from the data they are given. In modern post-training, that data is produced online by the model itself through \emph{rollouts}, and rollout design often dominates both the cost of training and the quality of the resulting learning signal.

A \emph{rollout} is a sampled trajectory from a prompt to termination. In text-only settings, a rollout reduces to a completion containing intermediate reasoning and a final answer; in tool- or environment-interactive settings, it includes action--observation loops and external feedback. Reasoning tasks make rollout design particularly consequential: the solution space is sparse and sensitive to early errors, and a single early mistake can invalidate an entire derivation or program. As a result, the rollout mechanism is not merely a decoding choice---it acts as a data-generation policy that determines what evidence is collected, which candidates survive, and where compute is spent. Common rollout design axes include topology, sampling diversity, granularity, stopping criteria, branching, and artifact reuse.

Recent progress makes these dependencies explicit. Process-level evaluators such as process reward models require step-delimited rollouts \citep{lightman2023let,zhao2025genprm,zou2025reasonfluxprm}; tree-structured rollouts rely on branching and backup rules \citep{yao2023tree,hou2025treerlllmreinforcementlearning,wu2025deepsearch}; and adaptive budgeting policies trade off rollout compute against accuracy by allocating more samples to informative prompts and skipping uninformative ones \citep{zheng2025actpaysefficientreinforcement,nguyen2026adaptiverolloutallocationonline}. Replay and recomposition further complicate the picture by reusing cached responses, recombining previously generated trajectories, or mining self-generated tasks across iterations without weight updates \citep{li2025repo,zhang2025improvingsamplingefficiencyrlvr,li2025rorecomp,chen2025selfevolvingcurriculumllmreasoning}. These design choices shape not only performance but also key desiderata such as reliability of supervision, coverage and informativeness of sampled trajectories, and sensitivity to compute budgets.

Despite their centrality, rollout strategies are often treated as implementation details and underreported. Existing surveys typically organize the space around optimization algorithms, reward modeling, or pipeline-wide categorizations of where RL appears \citep{kaufmann2025survey,wang2024reinforcement,srivastava2025technical,guo2025survey}, while reasoning- and agent-centric surveys focus on multi-step deliberation and search rather than the rollout pipeline itself \citep{zhang2025landscape,zhang2025survey}. This leaves rollout design implicit and obscures the link between data collection and learning outcomes. The resulting reproducibility challenge is significant: unreported changes in rollout configuration can confound comparisons and make it difficult to attribute improvements to the optimizer versus the data-generation pipeline.

This survey centers rollout design as a first-class object. We formalize rollout pipelines with unified notation and introduce \textbf{Generate--Filter--Control--Replay (GFCR)}, an optimizer-agnostic, lifecycle decomposition of rollout pipelines into four modular and composable stages---functionally distinct yet frequently interleaved, since filter signals trigger control decisions, replay artifacts seed future generation, and control policies determine what enters replay. \textbf{Generate} specifies how candidate trajectories are proposed (topology, scaffolding, sampling). \textbf{Filter} maps rollouts and prefixes to intermediate signals and optimizer-facing supervision (verifiers, judges, process scorers, learning-value diagnostics). \textbf{Control} allocates compute and makes continuation/branching/stopping decisions under budgets, and encompasses systems-level throughput optimization (speculative decoding, scheduling, and load balancing for rollout generation). \textbf{Replay} retains and reuses artifacts across rollouts without weight updates (buffers, caching, recomposition), and further encompasses self-evolving curricula in which rollouts autonomously generate new tasks, solutions, or agents that feed back into training. We complement GFCR with a \emph{criterion taxonomy} of reliability, coverage and informativeness, and cost sensitivity that characterizes the trade-offs rollout designs must navigate and provides a principled basis for evaluating design choices across modules. Figure~\ref{fig:framework-intro} provides an overview of the GFCR lifecycle and its interfaces.

\paragraph{Contributions.}
To our knowledge, this is the first survey that systematically organizes \emph{rollout strategies} for RL-based post-training of reasoning LLMs. Our primary contributions are:
\begin{itemize}
    \item We introduce GFCR, a unified taxonomy for describing rollout pipelines independently of the underlying optimizer, paired with a complementary criterion taxonomy (reliability, coverage, cost sensitivity) that provides a common vocabulary for comparing trajectory proposal, filtering, compute control, and reuse.
    \item We synthesize rollout-centric methods spanning RL with verifiable rewards, process supervision, judge-based gating, guided and tree/segment rollouts, adaptive compute allocation, early-exit and partial rollouts, systems-level throughput optimization, and replay/recomposition/self-evolution for self-improvement.
    \item We ground the taxonomy with case studies in mathematics, code/SQL, multimodal reasoning, tool-using agents, and agentic skill benchmarks---including skill induction, library management, and cross-task transfer---illustrating how interface constraints and feedback availability shape rollout design.
    \item We provide a structured diagnostic index that maps common rollout pathologies to GFCR modules and concrete mitigation levers, enabling practitioner-oriented troubleshooting.
    \item We identify open challenges, including verifier/judge calibration, principled compute accounting, and safe self-evolution with provenance tracking, and offer recommendations for reporting and evaluation to improve reproducibility.
\end{itemize}

\paragraph{Organization.}
We introduce global notation and formalize rollout pipelines in \S\ref{sec:problem}. We then survey the design space module by module: Generate (\S\ref{sec:g}), Filter (\S\ref{sec:f}), Control (\S\ref{sec:c}), and Replay (\S\ref{sec:r}), followed by domain case studies, practitioner-oriented diagnostics, and open problems.

%% file: latex/2_related.tex
\section{Related Work}
\label{sec:related}

\subsection{Comparison to Prior Surveys}

Recent progress in reasoning-focused post-training and inference time deliberation increasingly depends on how systems sample, structure, score, and reuse trajectories under a compute budget, including grouped sampling, branching search, tool interaction trajectories, stepwise versus terminal scoring, adaptive stopping, and replay. Most prior surveys organize the space primarily around feedback modeling, reward learning, and optimization objectives, leaving rollout strategy implicit \citep{kaufmann2025survey,jiang2025survey,chaudhari2025rlhf_deciphered}.

Surveys of RLHF and preference learning emphasize feedback collection and modeling, alignment loops, and evaluation protocols \citep{kaufmann2025survey,jiang2025survey,chaudhari2025rlhf_deciphered,ni2026survey,ni2025large}. Surveys of RL-enhanced LLMs summarize RLHF, RLAIF, and direct preference families and discuss optimization challenges \citep{wang2024reinforcement}. Technical surveys focus on RL algorithms and training mechanics for LLM fine-tuning \citep{srivastava2025technical,lireview}, while pipeline-wide surveys categorize where RL appears across data generation, pretraining, post-training, and test-time inference \citep{guo2025survey,hu2026figure,xia2025selection,xiesurvey,wu2024personalized}. Reasoning and agent-centric surveys review multi-step deliberation, search, and environment interaction \citep{zhang2025survey,zhang2025landscape,nguyen2025gui,wu2024visual}. In contrast, we center rollout strategy as the unit of analysis and provide GFCR as a modular vocabulary for comparing how topology, sampling, scoring granularity, budget allocation, and experience reuse compose into end-to-end systems. Accordingly, we focus on methods where rollout design plays a substantive role, rather than surveying RL algorithms or reward modeling in isolation.

%% file: latex/3_taxonomy_new.tex
\section{Foundations: Rollouts, Criteria, and the GFCR Framework} \label{sec:problem}

Modern LLM post-training increasingly relies on \emph{rollout-centric} pipelines: for a given prompt, a system generates one or more
candidate trajectories, evaluates them with verifiers or judges, and converts the resulting signals into training supervision.
In this setting, a \emph{rollout strategy} is rarely a single algorithmic knob---it is an end-to-end pipeline spanning how trajectories
are proposed, how they are scored, how compute is allocated under a budget, and what artifacts are retained and reused.
Without a shared vocabulary and notation, it becomes difficult to compare methods across domains (math/code/agents) or to isolate
which component is responsible for a reported gain.

This section establishes the foundations used throughout the survey. We first introduce \textbf{GFCR}, a functional decomposition of
rollout pipelines into \textbf{Generate}, \textbf{Filter}, \textbf{Control}, and \textbf{Replay} modules (Figure~\ref{fig:framework-gfcr}).
We then preview the GFCR taxonomy at a glance (Table~\ref{tab:gfcr_preview}), define global notation for rollouts, prefixes, and group sampling,
and conclude with a complementary \emph{criterion taxonomy} that summarizes the reliability/coverage/cost desiderata that rollout designs must trade off.

\subsection{GFCR as a Unifying Rollout Framework}
\begin{figure}[ht]
    \centering
    \includegraphics[width=1. \linewidth]{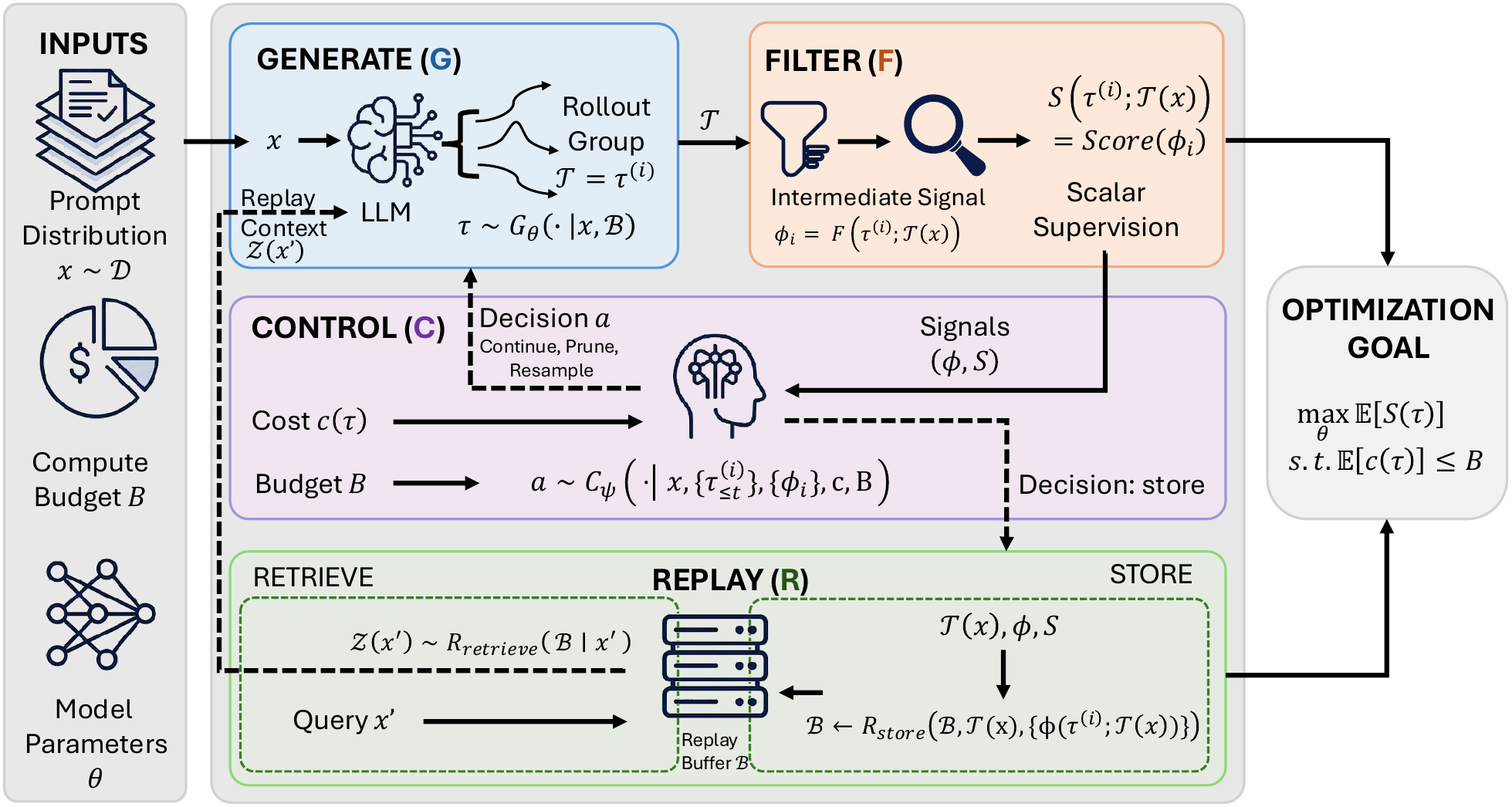}
    \caption{\textbf{GFCR as an end-to-end rollout system.} Given prompts $x\sim\mathcal D$ and a compute budget $B$, \textbf{Generate} samples a rollout group $\mathcal T(x)$; \textbf{Filter} maps each rollout to intermediate signals $\phi$ and training-facing supervision $S$; \textbf{Control} uses costs and signals to adapt continuation/pruning/resampling and decide what to store; and \textbf{Replay} retrieves/stores artifacts that condition future generation. The overall objective is to maximize expected utility $\mathbb{E}[S(\tau)]$ under compute constraints.}
    \label{fig:framework-gfcr}
\end{figure}

We view modern rollout pipelines for reasoning LLMs as compositions of four lifecycle modules that operate on rollouts (or
rollout prefixes) under compute budgets: \textbf{Generate (G)} proposes candidate trajectories; \textbf{Filter (F)} extracts
signals and converts them into training-facing supervision; \textbf{Control (C)} allocates compute and makes continuation,
branching, and stopping decisions; and \textbf{Replay (R)} retains and reuses artifacts across rollouts without weight
updates. Figure~\ref{fig:framework-gfcr} illustrates how these components compose into an end-to-end rollout system.

Crucially, these modules are often \emph{interleaved} rather than sequential: Filter signals can trigger Control decisions
such as pruning, early stopping, or adaptive resampling; Replay artifacts (cached responses, verified sub-traces, or mined
tasks) can seed future Generate calls; and Control policies determine which artifacts enter Replay and when replayed
artifacts are trusted, refreshed, or discarded. This decomposition provides a unified vocabulary for describing rollout
pipelines as modular design choices, rather than monolithic \emph{rollout strategies}.

To preview the remainder of the survey, Table~\ref{tab:gfcr_preview} summarizes our rollout-centric taxonomy, organized as four
composable lifecycle modules and indexed to the detailed sections.

\input{figures/gfcr_tree}

\subsection{Global Notation} \label{sec:notations}

\input{tables/notations}

We formalize \emph{rollouts} as the primary object manipulated by GFCR. Table~\ref{tab:notation} summarizes the global notation used throughout the paper (module-specific symbols are introduced in the corresponding GFCR sections).
Let $x \sim \mathcal{D}$ be a prompt/task input. A reasoning LLM with parameters $\theta$ induces a stochastic policy
$\pi_\theta$ that generates an interaction trajectory until termination. We write a rollout as
\begin{equation}
\tau = (x, u_{1:T}, o_{1:T}),
\end{equation}
where $u_t$ denotes the model’s output/action at step $t$ (text tokens and/or a tool/action call), $o_t$ is the
corresponding observation (tool output/environment response; empty in pure-text settings), and $T$ is a stopping time
(EOS, max length, success, environment done). Equivalently, define the (implicit) state as
$s_t=(x,u_{1:t-1},o_{1:t-1})$, sample actions $u_t \sim \pi_\theta(\cdot \mid s_t)$, and sample observations
$o_t \sim P(\cdot \mid s_t, u_t)$; this induces a trajectory distribution $p_\theta(\tau \mid x)$.

Many rollout pipelines also operate on \emph{prefixes} (partial rollouts), e.g., for step/segment scoring, pruning, or tree
expansion; we denote a prefix by $\tau_{\le t}=(x,u_{1:t},o_{1:t})$. In text-only settings, observations are empty, and a rollout reduces to a completion; we write $y \equiv u_{1:T}$ and use $\tau$ and $y$ interchangeably when environment interaction is irrelevant.
We also write $p_\theta(y\mid x)$ for the induced distribution over text-only completions.

A training system typically samples either a single rollout $\tau \sim p_\theta(\cdot \mid x)$ or a \emph{group} of $K$
rollouts
\begin{equation}
\mathcal{T}(x) = \{\tau^{(i)}\}_{i=1}^{K}, \qquad \tau^{(i)} \sim p_\theta(\cdot \mid x).
\end{equation}
Each rollout is assigned supervision that may depend on the full group (e.g., listwise judging, within-group baselines).
We denote intermediate \textbf{Filter} signals by $\phi(\tau^{(i)};\mathcal{T}(x))$ (validity, verification outcomes,
process scores, judge ranks, learning-value signals, etc.), and write the resulting training signal as
\begin{equation}
S(\tau^{(i)}; \mathcal{T}(x)) = \mathrm{Score}\!\left(\phi(\tau^{(i)}; \mathcal{T}(x))\right).
\end{equation}
For brevity, we often write $\phi_i \equiv \phi(\tau^{(i)};\mathcal{T}(x))$
and $S_i \equiv S(\tau^{(i)};\mathcal{T}(x))$ when the context is clear.
When the training signal depends only on a single trajectory (i.e., is not group-dependent), we write $S(\tau)$ as
shorthand.

We track rollout cost $c(\tau)$ (e.g., decoded tokens, tool calls, wall-clock) and impose budgets such as
$\sum_{i=1}^{K} c(\tau^{(i)}) \le B_x$. At a high level, rollout-centric RL aims to increase expected utility under compute
constraints:
\begin{equation}
\max_{\theta}\; \mathbb{E}_{x \sim \mathcal{D}}\; \mathbb{E}_{\tau \sim p_\theta(\cdot \mid x)}\!\left[S(\tau)\right]
\quad \text{s.t.}\quad \mathbb{E}[c(\tau)] \le B.
\end{equation}

Finally, real systems interleave generation, filtering, budget decisions, and reuse. We denote by
$q_{\theta,\mathrm{GFCR}}(\mathcal{T}\mid x,\mathcal{B})$ the distribution over rollout groups induced by the full GFCR
pipeline given prompt $x$ and current replay state $\mathcal{B}$. When the dependence on $\mathcal{B}$ is not central, we
may suppress it and write $q_{\theta,\mathrm{GFCR}}(\mathcal{T}\mid x)$. Detailed, module-specific formulations and
additional notation are introduced in the corresponding GFCR sections.

\paragraph{Pipeline-Induced Distributions and Module Interfaces.}
\label{sec:interfaces}
Recall from \S\ref{sec:notations} that $q_{\theta,\mathrm{GFCR}}(\mathcal{T}\mid x)$ denotes the distribution over rollout
groups induced by the full GFCR pipeline for a given prompt $x$. The i.i.d.\ sampling view
$\tau^{(i)} \sim p_\theta(\cdot\mid x)$ abstracts away the fact that modern systems are \emph{pipelines}: they interleave
generation, filtering, budgeted decision-making, and reuse across prompts and iterations.

Unlike $p_\theta(\tau\mid x)$, the induced distribution $q_{\theta,\mathrm{GFCR}}$ may depend on intermediate filter signals,
budgeting/stop decisions, adaptive sampling (e.g., variable $K$), and a persistent replay state $\mathcal{B}$ that stores
and reuses artifacts from past rollouts. Figure~\ref{fig:framework-gfcr} provides a schematic view of these interleavings. In
later sections, we instantiate each module (Generate/Filter/Control/Replay) with module-specific problem formulations and
notation, while retaining $q_{\theta,\mathrm{GFCR}}$ as the unifying object describing end-to-end rollout behavior.

\subsection{Rollout Criterion Taxonomy}
\label{sec:criteria}

A rollout strategy is defined not only by how it generates trajectories, but also by the criteria used to judge whether those
trajectories are useful for selection, learning, and compute allocation. We therefore organize rollout criteria along a small
set of high-level dimensions that capture what a pipeline is trying to optimize. This perspective is complementary to GFCR.
GFCR describes the functional modules that implement a rollout pipeline, whereas the criterion taxonomy describes the
desiderata that those modules must satisfy and the trade-offs that arise when they are optimized jointly.

\begin{figure}[ht]
    \centering
    \includegraphics[width=0.8\linewidth]{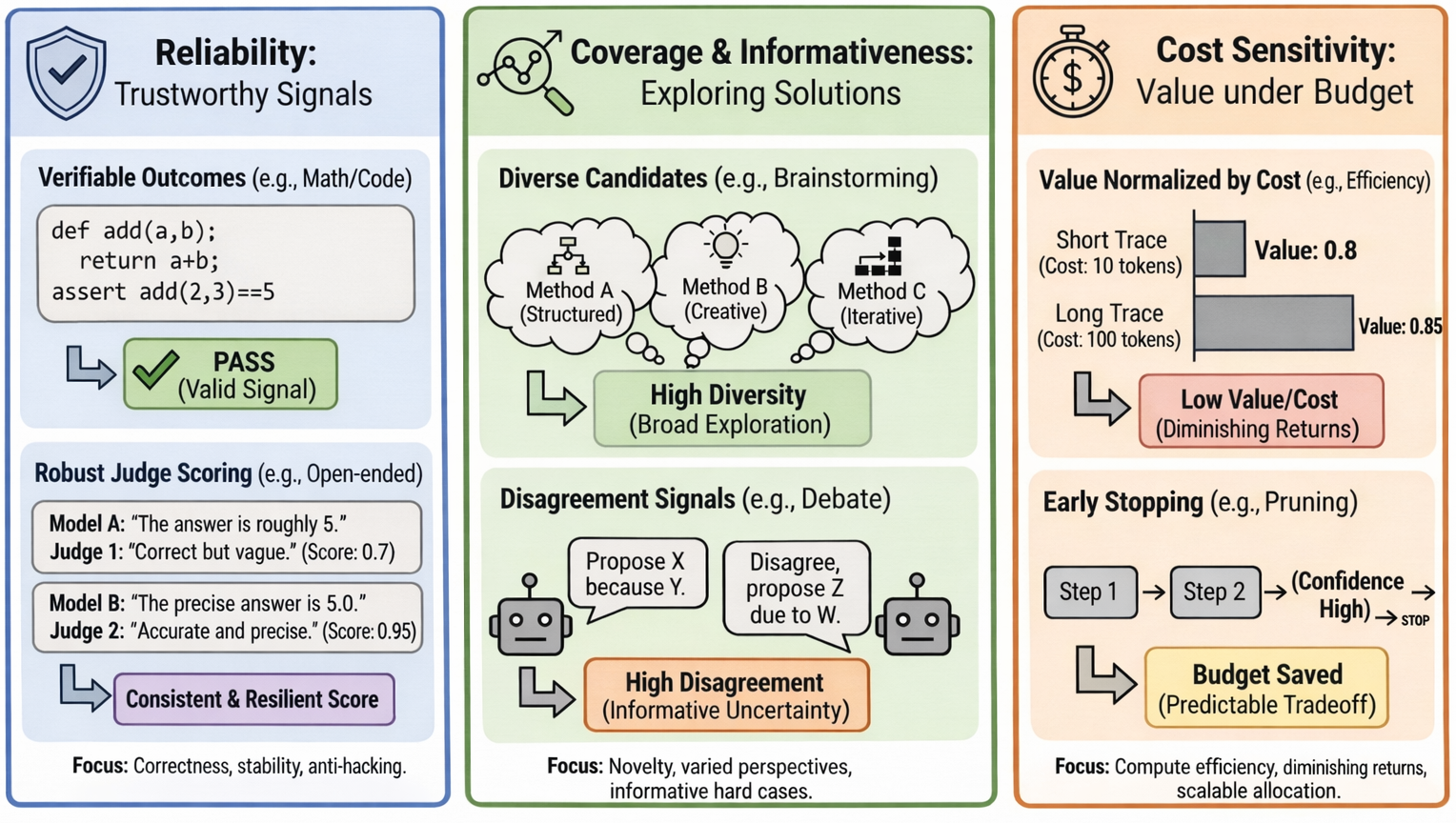}
\caption{\textbf{Rollout criterion taxonomy (at a glance).} We evaluate rollout strategies along three cross-cutting desiderata: \emph{reliability} (trustworthy signals via verifiers or robust judges), \emph{coverage \& informativeness} (diverse candidates and disagreement/uncertainty), and \emph{cost sensitivity} (value under compute budgets via value-per-cost and early stopping).}
    \label{fig:criterion}
\end{figure}

Concretely, GFCR is a \emph{functional decomposition}: it categorizes \emph{what} a rollout pipeline does (propose, measure,
decide, reuse). By contrast, the criterion taxonomy characterizes \emph{why} those choices are made and \emph{how} they should
be judged. For example, best-of-$K$ belongs in \textbf{Generate/Filter/Control}, while the question \emph{is best-of-$K$ worth it under
budget $B$ and does it improve reliability or merely increase variance?} belongs in the criterion taxonomy. Table~\ref{tab:rollout_criteria_refstyle_noaspect} summarizes the core rollout criteria we use throughout the survey to compare design choices across GFCR modules.~\Cref{fig:criterion} provides an at-a-glance visualization of the rollout criteria summarized in~\Cref{tab:rollout_criteria_refstyle_noaspect}.

\input{tables/criterion}

Reliability emphasizes that a rollout is only as useful as the trustworthiness of the signal derived from it. When tasks admit
automatic verification, reliability is dominated by correctness under verifiers. When verification is unavailable, reliability
depends on the stability and bias profile of model-based judges and on defenses against specification gaming. Coverage and
informativeness capture the fact that many pipelines benefit from sampling multiple candidates, not only to increase the chance
of success, but also to expose discriminative signals such as disagreement and near-miss behavior. Cost sensitivity captures
the compute-constrained nature of modern rollouts and motivates budget-aware policies that trade trajectory length, branching,
and sample count against utility, often using early stopping and reuse mechanisms.

%% file: figures/gfcr_tree.tex
\begin{table}[t]
\caption{GFCR taxonomy at a glance. Rollout behavior is assembled from four composable lifecycle modules.}
\centering
\small
\setlength{\tabcolsep}{6pt}
\renewcommand{\arraystretch}{1.15}
\label{tab:gfcr_preview}

\begin{tabularx}{\linewidth}{@{} l l Y @{}}
\toprule
\rowcolor{gfcrHead}
Module & Subcomponent & What it includes \\
\midrule

\gfcrsection{gfcrGbg}{Generate (G): How trajectories are proposed. \emph{Defines the proposal distribution + interaction structure.}}
\textbf{\S\ref{sec:g_topology}} & \textbf{Topology \& Interaction} &
Rollout structure and interaction patterns. \\
\textbf{\S\ref{sec:g_guidance}} & \textbf{Guidance \& scaffolding} &
Prompt structure, rubrics, and auxiliary traces. \\
\textbf{\S\ref{sec:g_sampling}} & \textbf{Sampling \& exploration} &
Decoding diversity and exploration signals. \\
\addlinespace[0.4em]

\gfcrsection{gfcrFbg}{Filter (F): Triage, scoring, and signal construction. \emph{Maps rollouts/prefixes to training-facing signals.}}
\textbf{\S\ref{sec:f_structural_validity}} & \textbf{Structural validity} &
Schema compliance, executability, and safety gates. \\
\textbf{\S\ref{sec:f-verification}} & \textbf{Correctness verification} &
Outcome checking via tests or solvers. \\
\textbf{\S\ref{sec:f-process}} & \textbf{Process quality} &
Step-level signals for reasoning quality. \\
\textbf{\S\ref{sec:f-comparative}} & \textbf{Comparative assessment} &
Ranking, judging, and calibration signals. \\
\textbf{\S\ref{sec:f-learningvalue}} & \textbf{Learning-value signals} &
Informativeness proxies: novelty and uncertainty. \\
\textbf{\S\ref{sec:f-supervision}} & \textbf{Training-signal construction} &
Scores mapped to labels or weights. \\
\addlinespace[0.4em]

\gfcrsection{gfcrCbg}{Control (C): Compute allocation and decision rules. \emph{Allocates compute; decides continue/stop under budgets.}}
\textbf{\S\ref{sec:c-task}} & \textbf{Prompt and task selection} &
Curricula and instance prioritization policies. \\
\textbf{\S\ref{sec:c-budget}} & \textbf{Budgeting \& scheduling} &
Adaptive compute allocation and scheduling. \\
\textbf{\S\ref{sec:c-config}} & \textbf{Rollout configuration control} &
Length, temperature, and efficiency constraints. \\
\textbf{\S\ref{sec:c-stop}} & \textbf{Continuation and stop rules} &
Early stopping and continuation criteria. \\
\textbf{\S\ref{sec:c-branch}} & \textbf{Branch and prune control} &
Branching policies and pruning heuristics. \\
\textbf{\S\ref{sec:c-offpolicy}} & \textbf{On-/off-policy controls} &
Mixing constraints and correction policies. \\
\textbf{\S\ref{sec:c-systems}} & \textbf{Systems considerations} &
Throughput optimization and systems constraints. \\
\addlinespace[0.4em]

\gfcrsection{gfcrRbg}{Replay (R): Retention, reuse, and self-evolution. \emph{Persists artifacts across rollouts without weight updates.}}
\textbf{\S\ref{sec:r-buffer}} & \textbf{Response resampling and retention} &
Buffers, caching, and experience reuse. \\
\textbf{\S\ref{sec:r-recompose}} & \textbf{Recomposition} &
Reusing verified segments to compose. \\
\textbf{\S\ref{sec:r-evolution}} & \textbf{Self-evolution} &
Generate new tasks via self-improvement. \\

\bottomrule
\end{tabularx}
\end{table}

%% file: tables/notations.tex
\begin{table}[t]
\caption{Global notation used throughout the paper. Module-specific notation is introduced in the corresponding GFCR sections.}    
\centering
\small
\setlength{\tabcolsep}{6pt}
\renewcommand{\arraystretch}{1.15}
\label{tab:notation}
\begin{tabularx}{\linewidth}{@{} l X @{}}
\toprule
Symbol & Meaning \\
\midrule
$\mathcal{D}$ & Distribution (dataset) over prompts/tasks. \\
$x$ & Prompt/task instance sampled from $\mathcal{D}$. \\
$\theta$ & Parameters of the language model / policy. \\
$\pi_\theta(\cdot \mid s)$ & Model policy over next output/action given state $s$. \\
$u_t$ & Model output/action at step $t$ (tokens and/or tool/action call). \\
$o_t$ & Observation at step $t$ (tool output/environment response; empty for text-only). \\
$T$ & Stopping time (EOS, max length, success, environment done). \\
$s_t=(x,u_{1:t-1},o_{1:t-1})$ & Implicit rollout state at step $t$. \\
$P(\cdot \mid s_t,u_t)$ & Environment/observation kernel (may be deterministic tools). \\
$\tau=(x,u_{1:T},o_{1:T})$ & A rollout trajectory. \\
$\tau_{\le t}$ & Prefix (partial rollout) up to step $t$. \\
$y$ & Text-only completion corresponding to a rollout (used when $o_{1:T}=\emptyset$). \\
$y^{(i)}$ & Text-only completion for the $i$-th rollout $\tau^{(i)}$ (i.e., $y^{(i)} \equiv u^{(i)}_{1:T}$). \\
$p_\theta(\tau \mid x)$ & Trajectory distribution induced by $\pi_\theta$ (and $P$). \\
$p_\theta(y \mid x)$ & Completion distribution induced by $\pi_\theta$ in text-only settings (i.e., the marginal of $p_\theta(\tau \mid x)$ when $o_{1:T}=\emptyset$). \\
$\mathcal{T}(x)=\{\tau^{(i)}\}_{i=1}^{K}$ & Group of $K$ rollouts sampled for the same prompt $x$. \\
$K$ & Number of rollouts in a group (may be adaptive). \\
$\phi(\tau;\mathcal{T})$ & Intermediate \textbf{Filter} signals computed from rollouts (validity, verification, process/judge scores, etc.). \\
$S(\tau;\mathcal{T})$ & Training signal derived from $\phi$ (may be group-dependent). \\
$t$ & Step index within a rollout ($t = 1,\ldots,T$). \\
$c(\tau)$ & Compute cost of a rollout (tokens, tool calls, wall-clock, etc.). \\
$B,\,B_x$ & Global budget / per-prompt budget constraint. \\
$\mathcal{B}$ & Replay buffer/cache storing past rollouts or sub-trajectories. \\
$q_{\theta,\mathrm{GFCR}}(\mathcal{T}\mid x,\mathcal{B})$ & Rollout-group distribution induced by the full GFCR pipeline given replay state $\mathcal{B}$ (we may suppress $\mathcal{B}$ when clear). \\
$t_{\mathrm{store}}$ & Timestamp/age associated with a replay entry (when stored in $\mathcal{B}$). \\
$\phi_i$ & Shorthand for $\phi(\tau^{(i)};\mathcal{T}(x))$ when context is clear. \\
$S_i$ & Shorthand for $S(\tau^{(i)};\mathcal{T}(x))$. \\
$G,F,C,R$ & \textbf{G}enerate / \textbf{F}ilter / \textbf{C}ontrol / \textbf{R}eplay modules in GFCR. \\
\bottomrule
\end{tabularx}
\end{table}

%% file: tables/criterion.tex
\newcommand{\hboldline}{\specialrule{1.0pt}{0pt}{0pt}}
\newcommand{\hsectionline}{\specialrule{0.8pt}{0pt}{0pt}}
\newcommand{\boldbottomline}{\specialrule{1.0pt}{0pt}{0pt}}

\begin{table}[!ht]
\centering
\caption{\textbf{Rollout criterion taxonomy (reference style).} We summarize three cross-cutting desiderata for rollout strategies---\emph{reliability}, \emph{coverage \& informativeness}, and \emph{cost sensitivity}---and list concrete signals and mechanisms that operationalize each criterion.} 
\label{tab:rollout_criteria_refstyle_noaspect}
\footnotesize
\renewcommand{\arraystretch}{1.10}
\setlength{\tabcolsep}{6pt}

\begin{tabularx}{\linewidth}{@{} l X @{}}
\toprule
\rowcolor{gfcrHead}
\textbf{Criterion} & \textbf{Description and Examples} \\
\hboldline

\rowcolor{gfcrF}
\multicolumn{2}{@{}l@{}}{\textsc{\textbf{Reliability}}} \\

\rowcolor{gfcrF!70}
\quad \textbf{Verifiable Outcomes} &
Executable / checkable rollouts yielding trustworthy supervision when verification exists (e.g., math or code): structural validity/executability gates, unit tests, symbolic checks, exact-match constraints. \\

\rowcolor{gfcrF!55}
\quad \textbf{Robust Judge Scoring} &
Stable evaluation when direct verification is unavailable (e.g., open-ended tasks): calibrated judging, consistency across prompt variants, and resistance to reward hacking. \\

\hline

\rowcolor{gfcrG}
\multicolumn{2}{@{}l@{}}{\textsc{\textbf{Coverage \& Informativeness}}} \\

\rowcolor{gfcrG!70}
\quad \textbf{Diverse Candidates} &
Broad exploration over plausible solution strategies (structured, creative, iterative), avoiding near-duplicate rollouts and encouraging novelty. \\

\rowcolor{gfcrG!55}
\quad \textbf{Disagreement Signals} &
Informative uncertainty revealing hard-but-solvable cases: disagreement across rollouts or critics, conflicting rationales, and high-variance outcomes. \\

\hline

\rowcolor{gfcrC}
\multicolumn{2}{@{}l@{}}{\textsc{\textbf{Cost Sensitivity}}} \\

\rowcolor{gfcrC!70}
\quad \textbf{Value Normalized by Cost} &
High utility per token/compute, exhibiting diminishing returns for longer traces or larger sample pools, with predictable quality--cost trade-offs. \\

\rowcolor{gfcrC!55}
\quad \textbf{Early Stopping (Pruning)} &
Adaptive truncation when confidence is high or marginal gains are low, enabling compute-efficient rollout allocation. \\

\boldbottomline
\end{tabularx}
\end{table}

%% file: latex/4.1_generate.tex
\section{Generate: How Trajectories Are Proposed}\label{sec:g}

\subsection{Problem Formulation: Generate (G)}
\label{sec:g-formulation}

\begin{figure}[ht]
    \centering
    \includegraphics[width=1.\linewidth]{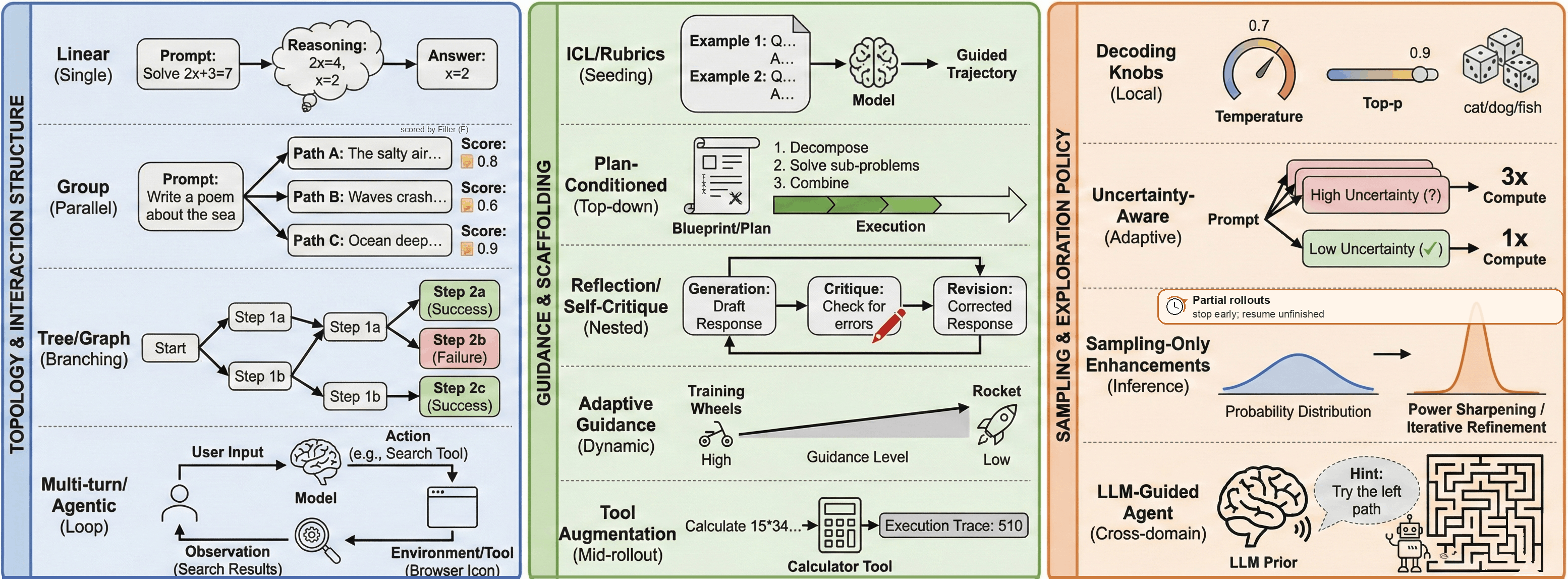}
\caption{\textbf{Generate module design space.} Rollout proposal mechanisms can be organized along three axes: \emph{(left)} topology \& interaction (single, group, tree/graph, and multi-turn/tool rollouts), \emph{(middle)} guidance/scaffolding $z$ (ICL/rubrics, plans, reflection, adaptive guidance, tool augmentation), and \emph{(right)} sampling \& exploration configuration $\kappa_G$ (decoding knobs, uncertainty-aware allocation, partial rollouts with resumption, and sampling-only inference enhancements). Group candidates are typically \emph{scored downstream} by Filter, and compute allocation may be coupled to Control.}
    \label{fig:framework-generate}
\end{figure}

Using the global notation in \S\ref{sec:notations}, the \textbf{Generate} module specifies how candidate rollouts are
proposed for a prompt $x \sim \mathcal{D}$. Its output is a candidate set (or structured collection) of rollouts
$\mathcal{T}(x)=\{\tau^{(i)}\}_{i=1}^{K}$ that becomes the input to \textbf{Filter} and is further shaped by \textbf{Control}
decisions. We model Generate as inducing a proposal distribution over candidate sets,
\begin{equation}
\mathcal{T} \sim q^{G}_{\theta}\!\left(\cdot \mid x;\; z,\,\kappa_G,\,\mathsf{Topo}\right),
\label{eq:g-proposal}
\end{equation}

where $q^{G}_{\theta}(\mathcal{T}\mid x;\, z,\kappa_G,\mathsf{Topo})$ denotes the Generate-module proposal distribution
over rollout groups $\mathcal{T}$ (induced by the current policy/model parameters $\theta$ under the specified rollout
procedure), $z$ denotes guidance/scaffolding information available at rollout time (e.g., rubrics, plans, critiques/repairs,
tool traces, retrieved exemplars), $\kappa_G$ denotes sampling/exploration configuration (e.g., temperature/top-$p$/diversity),
and $\mathsf{Topo}$ specifies the rollout topology and interaction pattern. Figure~\ref{fig:framework-generate} provides a visual map of these design axes (cf.\ Table~\ref{tab:generate_design_space_tradeoffs}).

In the simplest \emph{group} setting, the Generate module produces $K$ candidates that are
(approximately) conditionally independent given the rollout context and sampling configuration:
\begin{equation}
\mathcal{T}(x)=\{\tau^{(i)}\}_{i=1}^{K}, \qquad \tau^{(i)} \sim p_{\theta,\kappa_G}(\cdot \mid x;\, z),
\end{equation}
with $K=1$ recovering a single (linear) rollout.
Here $p_{\theta,\kappa_G}(\tau\mid x;\,z)$ denotes the trajectory distribution induced by running $\pi_\theta$
with decoding/sampling configuration $\kappa_G$ and guidance $z$ included in the rollout context/state;
in tool/environment settings, stochastic observations/transitions are drawn via $P(\cdot\mid s_t,u_t)$.

More structured topologies induce dependencies among candidates. For tree/graph rollouts, Generate expands partial
histories/prefixes $\tau_{\le t}$, producing a structured object (a search tree/graph) whose terminal nodes (leaves, in the
tree case) correspond to complete rollouts; shared prefixes amortize computation while enabling branching at uncertain points.
For tool/environment rollouts, $\tau$ includes both actions and observations, and the induced rollout distribution depends on
the environment kernel $P(\cdot\mid s_t,u_t)$ (transitions/observations). Across all cases, topology ($\mathsf{Topo}$),
scaffolding ($z$), and sampling knobs ($\kappa_G$) jointly determine the diversity, support, and cost profile of the proposed
candidates that the rest of the GFCR pipeline operates on.

\input{figures/generate_design}

Table~\ref{tab:generate_design_space_tradeoffs} summarizes the primary trade-offs across topology/interaction, guidance/scaffolding, and sampling/exploration choices used in Generate.

\subsection{Topology and Interaction }\label{sec:g_topology}

The topology of a rollout determines how the policy explores the solution space and how feedback signals are distributed across generated traces. We identify four principal topological categories that span current practice. \textbf{Linear rollouts} sample one complete trajectory per prompt (per update), typically applying an outcome reward once the sequence terminates; this is the dominant paradigm in early code and reasoning RL. \textbf{Group rollouts} sample $K$ independent candidates in parallel from the same prompt, enabling within-group comparisons, majority voting, and relative-advantage estimation, which underpins GRPO-style methods. \textbf{Tree and graph rollouts} branch at intermediate prefixes, attach verifier or partial signals to internal nodes, and allocate rollout budget via expansion and pruning. Finally, \textbf{multi-turn and agentic rollouts} instantiate an action\textendash{}observation loop, interleaving model actions with environment observations (tool outputs, API responses, execution traces, or simulator feedback) and requiring the policy to reason over stateful, multi-step interaction.

\paragraph{Linear Single-Trajectory Rollouts.}

Linear rollouts generate a single trajectory per prompt, representing the simplest topology and the inherited default from classic policy gradient methods. In this setting, the policy samples one complete response, receives a terminal reward, and updates without within-prompt comparisons. Variance reduction becomes critical since no peer responses exist to form relative baselines.

In code generation, CodeRL~\citep{le2022coderl} trains an actor-critic framework where a critic network predicts functional correctness and provides dense feedback for the actor. RLTF~\citep{liu2023rltf} extends this approach with an online generation framework and multi-granularity unit test feedback, providing finer credit assignment than coarse episode-level rewards. These examples illustrate how linear rollouts can still produce useful training signals when coupled with richer evaluators and feedback extraction.
Separately, treating chain-of-thought as latent state transitions yields structured rollout dynamics and exploration-aware proposal distributions rather than purely token-level heuristics~\citep{wuctrls}.

\paragraph{Group Rollouts (Parallel Candidates).}

Group rollouts sample $K$ independent candidates per prompt, enabling within-prompt variance reduction without a learned critic. Rather than comparing a single response against a value estimate, the model uses the group itself as a reference distribution, computing advantages through normalization: $\hat{A}_i = (r_i - \mu_G) / \sigma_G$, where $\mu_G$, $\sigma_G$ are the mean and std of rewards within the group.

GRPO~\citep{shao2024deepseekmath} introduced this paradigm, and DeepSeek-R1~\citep{guo2025deepseek} demonstrated that applying it directly to base models can elicit emergent reasoning behaviors. Practical extensions address common failure modes. DAPO introduces \emph{decoupled clipping} and a token-level policy-gradient loss, and uses \emph{dynamic sampling} to avoid zero-gradient updates when within-group rewards are uniform (all correct or all incorrect)~\citep{yu2025dapo}. Length inflation, where models produce increasingly verbose responses, is controlled by filtering on response length or reward-per-token ratio~\citep{shrivastava2025samplethinklessgroup}. Dense step-wise rewards extend the paradigm to credit intermediate reasoning steps in multimodal settings~\citep{zhang2025r1}.

Many recent methods retain prompt-level group sampling while modifying rollout curation, exploration, or learning signals, including GRPO-Lead \citep{zhang2025grpo}, Not All Rollouts are Useful \citep{xu2025rolloutsusefuldownsamplingrollouts}, XRPO \citep{bamba2025xrpo}, G$^2$RPO-A \citep{guo2025g}, Scaf-GRPO \citep{zhang2025scaf}, SEED-GRPO \citep{chen2025seedgrposemanticentropyenhanced}, and Stepwise Guided Policy Optimization (SGPO) \citep{chen2025stepwise}.
POPE~\citep{huang2025pluralistic} rollouts diverse responses to enable pluralistic preferences and enable better coverage of the LLM policy.
ADPO~\citep{huang2025image} leverages image pairs as comparative prompting for induced image difference caption rollouts.
Related formulations for group rollouts focus on how the \emph{within-group baseline} is computed. For example, leave-one-out baselines (RLOO) use the other samples in the same prompt-level group as a reference signal \citep{ahmadian2024back}. On the analysis side, recent theoretical work characterizes GRPO's effective loss/dynamics for binary verifiable rewards and discusses its success-amplification behavior \citep{mroueh2025reinforcementlearningverifiablerewards}.
\paragraph{Tree/Graph Rollouts (Branching Search).}\label{sec:g_topology_tree}

Tree/graph rollouts generalize group sampling by branching at intermediate prefixes and reusing shared prefixes as internal nodes. Leaves receive outcome or verifier rewards, and these signals are backed up to internal nodes to obtain step- or segment-level supervision and to allocate rollout budget via expansion and pruning. TreeRPO, for example, uses tree sampling and bottom-up reward propagation to estimate step-level rewards without a separate process reward model, then computes advantages over step-level groups within the tree~\citep{yang2025treerpo}. TreeRL couples on-policy tree search with intermediate process supervision to allocate exploration across branches under a fixed token budget~\citep{hou2025treerlllmreinforcementlearning}.
Additional tree/graph rollout variants for verifiable-reward post-training include heuristic tree modeling for policy optimization and inference \citep{li2025treepo}, MCTS-style branching and backup to overcome RLVR bottlenecks \citep{wu2025deepsearch}, off-policy tree-guided advantage optimization \citep{huang2025treeopo}, and tree-search formulations for agent RL \citep{ji2025treegrpo}. Related designs study tree-structured credit assignment \citep{tran2025tempo}, lookahead tree rollouts for trajectory-level exploration \citep{xing2025latr}, and tool-use training with rewarded trees \citep{wu2025portool}, while some approaches bridge SFT and RL via branched rollouts from expert anchors \citep{zhang2025bread}. 
Listwise preference diffusion optimization~\citep{huanglistwise} generates diverse user-behavior trajectories via diffusion decoders, improving batch diversity in multi-step rollouts.
Analogous branching rollouts also appear in diffusion/flow post-training \citep{li2025branchgrpo,ding2025treegrpo,dynamictreerpo2025}. Tree/graph rollouts shift compute from independent full trajectories to \emph{structured reuse}: shared prefixes amortize KV-cache and reduce redundant computation, while branching concentrates exploration on uncertain steps. The trade-off is higher control complexity (expansion/pruning policies, node bookkeeping) and additional design choices for how to back up sparse leaf outcomes into dense process credit.

\paragraph{Multi-Turn, Tool-Using, and Multi-Agent Rollouts.}
Multi-turn rollouts generalize single completions by placing the policy in an action\textendash{}observation loop: at each step the model emits an action (tool call, edit, navigation, message) and conditions on the resulting observation (tool output, execution trace, retrieved context), with state carried across turns. Learning signals are often sparse and delayed, so practical methods either (i) rely on environments with verifiable endpoints such as unit tests or task success, or (ii) construct denser per-step credit from interaction traces. SWE-Gym operationalizes software engineering as an executable environment and trains agents and verifiers on collected trajectories \citep{pan2024training}, while WebRL targets web interaction and uses online curriculum RL to cope with sparse feedback in browser tasks \citep{qi2024webrl}; related work trains repository-level agents with a minimal tool API (e.g., a jump-to-definition tool) to navigate codebases end-to-end with RL \citep{zhang2025one}. 
SceneAlign~\citep{wang2026scenealign} leverages an intermediate scene graph strucutre for more visually grounded rollouts.
CSyMR~\citep{wang2025csymr} integrates Music21 toolkit for better symbolic music understanding generation.
On the optimization side, GiGPO extends group-based objectives to long-horizon interaction via \emph{anchor-state grouping} (identifying repeated states across trajectories) to form step-level groups and provide localized credit without a critic \citep{feng2025gigpo}. 
RAGEN \citep{wang2025ragen} studies multi-turn RL dynamics in LLM agents, Search-R1 \citep{jinsearch} introduces multi-turn rollout with a search engine, and VerlTool \citep{jiang2025verltool} provides a tool-use interface for agentic RL pipelines. 
SAND~\citep{xia2025sand} enables LLM agents to explicitly rollout over candidate actions before committing to one.
Doc-React~\citep{wu2025doc} instantiates iterative retrieve-and-read rollouts over long heterogeneous documents, using the LLM as a judge and generator to refine sub-queries under an information-gain objective.
Finally, multi-agent rollouts treat the observation stream as strategic: ARLAS frames indirect prompt injection defense as a two-player game with adversarial attackers \citep{wang2025adversarial}, and related work incorporates simulated users or self-play loops to generate interactive training experience \citep{zhao2025muarl,wei2025ssr}.

\subsection{Guidance and Scaffolding}\label{sec:g_guidance}

\paragraph{ICL Seeding and Structured Rubrics.}

Guidance can be injected at rollout start via ICL exemplars, reasoning templates, or explicit
evaluation rubrics, which biases the proposal distribution toward more structured trajectories.
This perspective naturally connects to LLM-as-a-Judge: MT-Bench highlights that judgment outcomes can depend strongly on the judge prompt design and evaluation protocol \citep{zheng2023judging}, while subsequent analyses identify systematic biases, such as position bias, that arise from prompt structure \citep{shi-etal-2025-judging}, motivating rubric-aware mitigation strategies summarized in recent surveys \citep{gu2024llmasajudge}.
Beyond evaluation, such seeding can also be used inside RL training to break \emph{all-failed}
regimes: XRPO~\citep{bamba2025xrpo} allocates rollout budget to ICL-augmented prompts by injecting curated exemplars (e.g., solved examples) when base rollouts yield no successes, explicitly aiming to break zero-reward symmetry.

\paragraph{Plan- or Answer-Conditioned Rollouts.}

Rollouts can be guided by conditioning on explicit targets, such as high-level plans, imposing
top-down constraints on trajectory generation. Separately, some methods introduce
reference-answer-guided signals during training that bias the rollout distribution toward desired outcomes
without requiring a fixed intermediate trace.
Such target-conditioned guidance provides a global objective signal that reduces search
ambiguity in long-horizon reasoning, complementing local or step-level guidance mechanisms.
Plan-then-act methods first generate a global plan and then execute reasoning steps under
this scaffold, using the plan to restrict and organize subsequent exploration \citep{dou2025plan},
while reference-answer-guided approaches use known answers (or answer-derived signals) during training to steer the distribution
over reasoning trajectories while allowing flexibility in intermediate reasoning \citep{lin2025ravr}.

\paragraph{Reflection and Self-Critique as Sub-Rollouts.}

Reflection-based guidance treats critique or repair not as post-processing, but as an
integral part of the rollout itself. In this view, reasoning is augmented with explicit
evaluation or revision stages that form nested sub-rollouts, each with its own acceptance
or scoring signal, allowing the model to reassess and refine intermediate trajectories
before finalization.
Critic-CoT~\citep{criticcot} exemplifies this pattern by introducing an explicit chain-of-thought critic that
evaluates and revises intermediate reasoning prior to producing a final answer.
Related work further shows that such self-generated critiques can serve as effective
training signals, improving reward modeling quality when incorporated into RL pipelines
\citep{criticrm}.
More generally, even in the absence of an explicit verifier, some methods derive rewards from model-internal signals. RLPR~\citep{yu2025rlpr}, for example, uses the model's intrinsic probability of reference-answer tokens to construct a verifier-free training signal.

\paragraph{Guided Rollouts with Adaptive Guidance Strength.}

Adaptive guidance views assistance as a dynamic control signal during rollouts: guidance is
increased when the policy stalls or collapses into zero-reward regimes, and gradually reduced
as competence improves. Rather than enforcing fixed expert trajectories, these methods aim to
provide minimal, temporary support that restores learning signals while preserving on-policy
exploration and autonomy.
G$^2$RPO-A \citep{guo2025g} instantiates this idea by adaptively modulating guidance strength within GRPO based on training dynamics, injecting assistance when needed and fading it as competence improves.
Scaf-GRPO \citep{zhang2025scaf} further formalizes adaptive guidance as a scaffolded curriculum, activating
hierarchical hints only for true-hard problems and fading them as the model internalizes the
required skills.
At a finer granularity, Stepwise Guided Policy Optimization~\citep{chen2025stepwise} targets the \emph{all-negative}
regime by using a step-wise judge model to incorporate response diversity within groups, enabling learning even when initial groups contain no correct samples .

\paragraph{Tool Augmentation and Execution Traces.}

Tool-augmented rollouts treat external tools as first-class components of the trajectory: tools are invoked mid-rollout for execution, retrieval, or analysis, and their inputs and
outputs become part of the reasoning trace. This expands the rollout space from pure text
generation to interactive trajectories grounded in executable environments, enabling more
reliable evaluation and learning signals. RLTF \citep{liu2023rltf} exemplifies this paradigm by incorporating unit test execution results into rollouts, using test feedback as a structured reward signal for reinforcement learning in code generation tasks. Similarly, benchmarks and evaluation suites such as BIRD \citep{li2023llmservedatabaseinterface} (SQL executed against databases) and SWE-Bench \citep{jimenez2023swe} (patches validated by running test harnesses) operationalize execution-grounded feedback, making execution traces and tool outputs natural filtering signals.

\subsection{Sampling and Exploration Policy}
\label{sec:g_sampling}

Sampling and decoding policies determine how the model explores the trajectory space, shaping both the diversity of candidate rollouts and the efficiency of the learning signals derived from them. Local decoding choices modulate token-level randomness, while uncertainty-aware sampling makes exploration input-adaptive by allocating compute preferentially to informative prompts. In addition, some methods modify the sampling policy without updating model weights, and others use LLM-generated hints to guide exploration in external environments, so we include them here because they directly influence the rollout distribution.

\paragraph{Decoding and Diversity Knobs.}
Stochastic decoding mechanisms, including temperature scaling, top-$p$ sampling, and diversity penalties, modulate the entropy of the token-level proposal distribution and thereby shape exploration in the trajectory space. Higher entropy encourages diversity, although it also increases variance and may produce excessively long or incoherent generations. This has motivated approaches that manage how diversity is realized during rollout generation, so that additional randomness contributes an informative exploratory signal rather than computational waste. APRIL~\citep{zhou2025aprilactivepartialrollouts} addresses inefficiency arising from long-tailed rollout length distributions by over-provisioning rollout requests and stopping generation once a target number of samples has completed. Unfinished rollouts are resumed in later iterations, which preserves data while implicitly reshaping the realized sampling distribution and improving throughput without degrading task performance. In parallel, TreeRL~\citep{hou2025treerlllmreinforcementlearning} reallocates exploration within a rollout by using on-policy tree search instead of independent chains, enabling branching under a fixed token budget. These methods indicate that diversity control in reinforcement learning for large language models involves not only token-level sampling parameters, but also the organization and allocation of exploratory behavior across rollouts.

\paragraph{Uncertainty-Aware Sampling and Exploration Signals.}
Uncertainty can be used during rollout generation to prioritize informative prompts and allocate sampling compute more effectively. In group-based policy gradient training, prompts whose responses are consistently correct or consistently incorrect contribute little gradient signal, while prompts near the model decision boundary tend to be more informative. Recent work therefore estimates uncertainty from reward outcomes or semantic disagreement and introduces it at different stages of the rollout pipeline.

One line of work uses prompt-level uncertainty to guide sample selection and curriculum scheduling, without explicitly modifying the number of rollouts per prompt. 
VCRL~\citep{jiang2025vcrl} treats reward variance as a proxy for difficulty and prioritises prompts with intermediate variance, which tend to lie near the model's decision boundary.
VADE~\citep{hu2025vade} models prompt difficulty with per-sample Beta distributions and uses Thompson sampling to select prompts that maximize an information-gain objective. 
MMR1~\citep{leng2025mmr1enhancingmultimodalreasoning} extends these ideas to multimodal RL, combining reward variance and trajectory diversity to stabilise GRPO training. 
Across these approaches, reward-side variance functions as a practical signal of epistemic uncertainty, encouraging learning on prompts that are difficult but not hopeless.

A complementary direction uses uncertainty to adapt the allocation of rollout compute itself, so that ambiguous prompts attract additional samples while easy or saturated ones are deprioritised. 
Adaptive rollout allocation can be framed as an expected-gradient-variance minimization problem under a fixed compute budget, assigning more samples to prompts expected to reduce estimator noise \citep{nguyen2026adaptiverolloutallocationonline}. 
XRPO~\citep{bamba2025xrpo} introduces targeted exploration/exploitation mechanisms, including ICL seeding for zero-reward prompts and an uncertainty-reduction-motivated rollout allocator under a fixed compute budget.
AR3PO~\citep{zhang2025improvingsamplingefficiencyrlvr} adds response reuse to this paradigm, leveraging previously generated correct responses while adaptively allocating additional samples to difficult prompts.

Finally, some work focuses less on the amount of sampling and more on the quality of the exploration signal. 
SEED-GRPO~\citep{chen2025seedgrposemanticentropyenhanced} measures semantic entropy, meaning the diversity of meanings expressed across candidate outputs, as an indicator of epistemic uncertainty, and uses it to modulate update magnitude to avoid over-updating on uncertain cases. 
Unlike reward variance, semantic entropy captures ambiguity in reasoning structure rather than in scalar outcomes, providing a complementary view of uncertainty.

\paragraph{Sampling-Only Reasoning Enhancements.}
Sampling-only methods aim to improve reasoning purely at inference time by modifying the sampling policy without updating model parameters. These approaches leverage additional compute rather than supervision and therefore serve as practical rollout primitives and baselines. Power Sampling method~\citep{karan2025reasoningsamplingbasemodel} approximates the distribution-sharpening effect of RL by sampling from a power distribution proportional to $p_\theta(y\mid x)^\alpha$ (for $\alpha>1$), using an MCMC procedure to accept or reject proposed continuations. This concentrates probability mass on higher-probability reasoning traces, yielding gains without training-time supervision.

\paragraph{LLM-Guided Exploration for RL Agents (Cross-Domain).}
In sparse-reward reinforcement-learning settings, large language models can supply semantic priors that bias exploration toward meaningful state–action regions rather than relying on uninformed stochastic search. Recent work operationalises this idea by conditioning rollouts on LLM-generated guidance while leaving policy optimization to the RL agent. ExploRLLM~\citep{ma2025explorllmguidingexplorationreinforcement} integrates foundation models by augmenting the agent's observation and by providing base and exploration policies, while the learned RL policy adapts by learning residual corrections. Complementarily, LLM-hints methods~\citep{jain2025guidingexplorationreinforcementlearning} inject LLM-generated action hints into the observation stream, enabling the policy to learn when to rely on or ignore external guidance. Across these approaches, semantically guided rollouts substantially improve exploration efficiency in sparse-reward domains while preserving policy autonomy.

%% file: figures/generate_design.tex
\begin{table}[ht]
\centering
\scriptsize
\setlength{\tabcolsep}{4pt}
\renewcommand{\arraystretch}{1.15}
\caption{\textbf{Generate design space and primary trade-offs.} Rows summarize rollout-construction choices used in representative methods, emphasizing the main efficiency/quality trade-off each choice introduces.}
\label{tab:generate_design_space_tradeoffs}
\begin{tabularx}{\linewidth}{@{} p{0.12\linewidth} p{0.19\linewidth} Y Y @{}}
\toprule
\rowcolor{gfcrHead}
\textbf{Subcomponent} & \textbf{Design choice} & \textbf{Primary trade-off} & \textbf{Representative works} \\
\midrule

\rowcolor{gfcrG}
\multicolumn{4}{@{}l@{}}{\textbf{Topology \& interaction ($\mathsf{Topo}$)}}\\
\rowcolor{gfcrG!70}
\textbf{Topology} & Linear (single rollout) &
Simplicity$\uparrow$; within-prompt baseline$\times$; gradient variance$\uparrow$. &
\citep{le2022coderl,liu2023rltf}. \\
\rowcolor{gfcrG!55}
\textbf{Topology} & Group ($K$ rollouts) &
Variance$\downarrow$ via within-prompt baselines; compute$\uparrow$ with $K$. &
\citep{shao2024deepseekmath,guo2025deepseek,yu2025dapo}. \\
\rowcolor{gfcrG!40}
\textbf{Topology} & Tree/graph (branching prefixes) &
Prefix reuse$\uparrow$ and targeted exploration; control/bookkeeping$\uparrow$. &
\citep{hou2025treerlllmreinforcementlearning,bamba2025xrpo}. \\
\rowcolor{gfcrG!50}
\textbf{Interaction} & Tool / environment loop &
Grounded feedback$\uparrow$; non-stationarity/latency$\uparrow$ (env-dependent). &
\citep{liu2023rltf,li2023llmservedatabaseinterface,jimenez2023swe}. \\
\hline
\rowcolor{gfcrF}
\multicolumn{4}{@{}l@{}}{\textbf{Guidance \& scaffolding ($z$)}}\\
\rowcolor{gfcrF!70}
\textbf{Seeding} & ICL exemplars / rubrics &
Early success and consistency$\uparrow$ on hard prompts; risk of rubric/format bias transfer. &
\citep{bamba2025xrpo,zheng2023judging,shi-etal-2025-judging}. \\
\rowcolor{gfcrF!55}
\textbf{Conditioning} & Plan-then-act / targets &
Long-horizon structure$\uparrow$; overhead/rigidity$\uparrow$. &
\citep{dou2025plan,lin2025ravr}. \\
\rowcolor{gfcrF!40}
\textbf{Nested rollouts} & Reflection / critique / repair &
Error recovery$\uparrow$; tokens/latency$\uparrow$. &
\citep{criticcot,criticrm}. \\
\rowcolor{gfcrF!50}
\textbf{Adaptive guidance} & Adaptive corrective guidance &
Stability$\uparrow$ under failure modes; extra guidance-generation/tuning overhead. &
\citep{guo2025g,zhang2025scaf,chen2025stepwise}. \\
\hline

\rowcolor{gfcrC}
\multicolumn{4}{@{}l@{}}{\textbf{Sampling \& exploration ($\kappa_G$)}}\\
\rowcolor{gfcrC!70}
\textbf{Sampling} & Group sampling and response filtering &
Signal quality$\uparrow$ (avoid zero-gradient/low-value samples) with higher train-time sampling cost. &
\citep{yu2025dapo,shrivastava2025samplethinklessgroup}. \\
\rowcolor{gfcrC!65}
\textbf{Partial rollouts} & Resume unfinished samples &
Throughput$\uparrow$ under heavy-tailed lengths; resumption/bookkeeping$\uparrow$. &
\citep{zhou2025aprilactivepartialrollouts}. \\
\rowcolor{gfcrC!55}
\textbf{Uncertainty signals} & Uncertainty-aware sampling / weighting &
Sample efficiency$\uparrow$ on boundary prompts; estimator overhead and potential tail under-training. &
\citep{jiang2025vcrl,hu2025vade,chen2025seedgrposemanticentropyenhanced}. \\
\rowcolor{gfcrC!50}
\textbf{Adaptive compute} & Adaptive $K$ / allocation / reuse &
Budget efficiency$\uparrow$ via per-prompt allocation and reuse; scheduler/buffer complexity$\uparrow$. &
\citep{nguyen2026adaptiverolloutallocationonline,zhang2025improvingsamplingefficiencyrlvr,bamba2025xrpo}. \\
\rowcolor{gfcrC!40}
\textbf{Sampling-only} & No-update baselines &
Test-time gains without training; no learning signal (no weight updates). &
\citep{karan2025reasoningsamplingbasemodel}. \\
\bottomrule
\end{tabularx}
\end{table}

%% file: latex/4.2_filter.tex
\section{Filter: From Rollouts to Learning Signals }

\label{sec:f}

\subsection{Problem Formulation: Filter (F) --- From Rollouts to Learning Signals}
\label{sec:f-formulation}

Using the global notation in \S\ref{sec:notations}, the \textbf{Filter} module maps sampled candidates
$\mathcal{T}(x)=\{\tau^{(i)}\}_{i=1}^{K}$ into intermediate signals and optimizer-facing supervision used for
pruning/selection, credit assignment, and compute allocation. In text-only settings we write $y^{(i)} \equiv u^{(i)}_{1:T}$
for the completion corresponding to $\tau^{(i)}$. Filter may operate on complete rollouts $\tau^{(i)}$ as well as prefixes
$\tau_{\le t}^{(i)}$ (e.g., step/segment-level scoring in long traces or during tree expansion).

Abstractly, Filter produces intermediate signals
\begin{equation}
\phi_i \;=\; F\!\big(\tau^{(i)};\mathcal{T}(x)\big),
\label{eq:f-phi}
\end{equation}
where $\phi_i$ may include validity indicators, verifier outcomes, process/step scores, judge preferences or ranks, and
learning-value diagnostics such as disagreement, entropy, or novelty.

A convenient decomposition separates filtering into (i) gating, (ii) semantic evaluation, and (iii) mapping into
training targets, possibly using the entire group:
\begin{align}
m_i &= \mathrm{Gate}(x, \tau^{(i)}) \in \{0,1\}, \label{eq:f-gate}\\
r_i^{\mathrm{raw}} &= \mathrm{Eval}(x, \tau^{(i)}) \in \mathbb{R}, \label{eq:f-eval}\\
(w_i,\,\tilde r_i,\,\tilde A_i,\,\ell_i) &= \mathrm{Map}\Big(\{(m_j,r_j^{\mathrm{raw}},\tau^{(j)})\}_{j=1}^{K}\Big). \label{eq:f-map}
\end{align}

In text-only settings, $\tau^{(i)}$ reduces to $y^{(i)} \equiv u^{(i)}_{1:T}$, and we may write $\mathrm{Gate}(x,y^{(i)})$ and $\mathrm{Eval}(x,y^{(i)})$.

\begin{figure}[ht]
    \centering
    \includegraphics[width=1.0\linewidth]{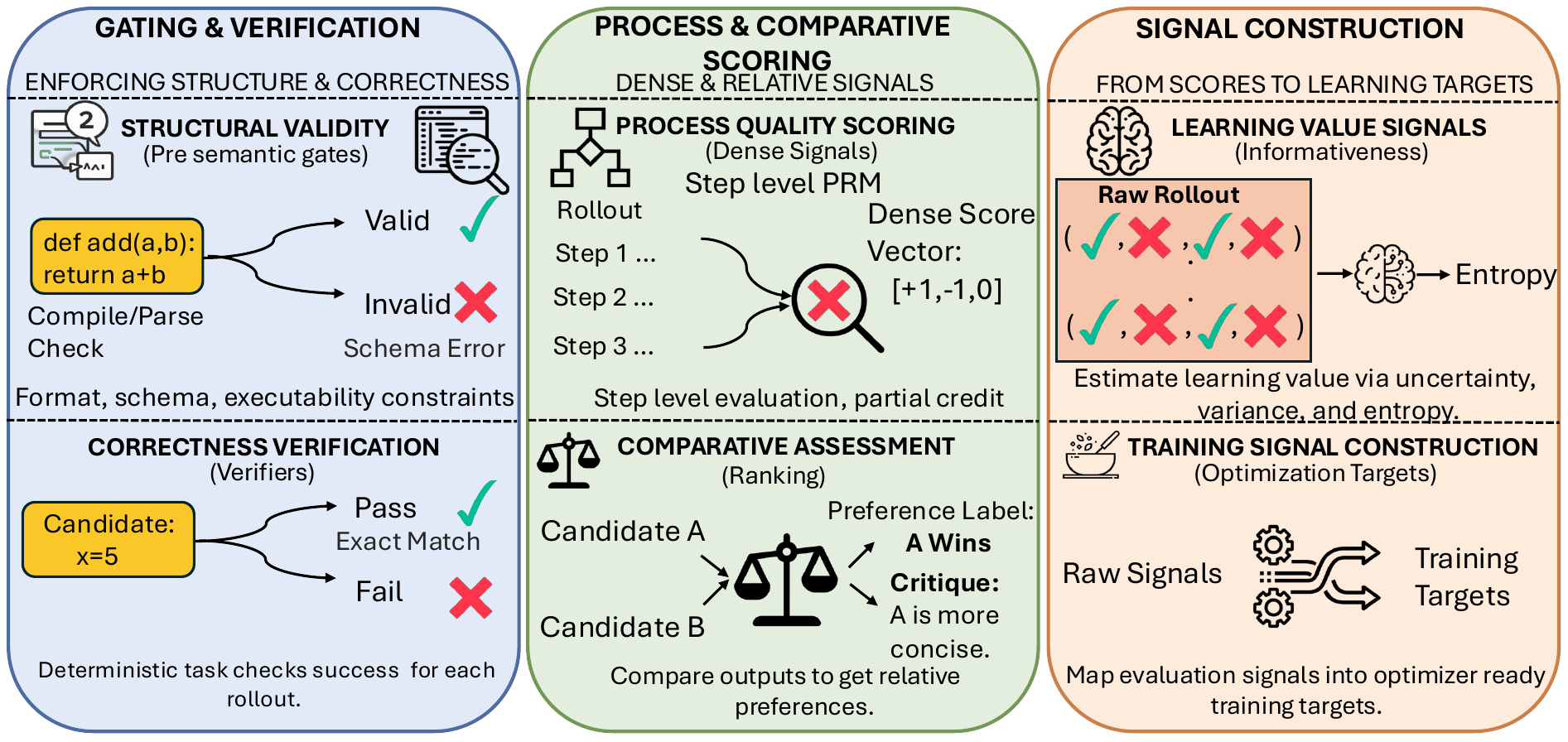}
    \caption{\textbf{Filter module design space.} Filter maps sampled rollouts (or rollout prefixes) into intermediate signals and optimizer-facing supervision. It can be organized along three components: \emph{(left)} gating \& verification (structural validity, schema/executability checks, deterministic correctness tests), \emph{(middle)} process \& comparative scoring (dense step-level process signals and relative judgments via ranking/judging), and \emph{(right)} signal construction (learning-value diagnostics such as uncertainty/variance/entropy, and mapping raw outcomes into optimizer-ready targets such as masks/weights, advantages, or preference labels). Filter outputs are consumed by the optimizer and may also inform Control decisions.}

    \label{fig:framework-filter}
\end{figure}

Here $m_i$ captures structural validity (format/schema/executability/safety), $r_i^{\mathrm{raw}}$ is a raw evaluation signal (e.g.,
deterministic verification correctness, judge score, or process-quality score), and $\mathrm{Map}(\cdot)$ converts raw
outcomes into optimizer-facing supervision: sample weights $w_i$ (masking/reweighting), shaped rewards $\tilde r_i$,
advantages $\tilde A_i$, or discrete labels/preferences $\ell_i$ (pairwise or listwise). Group-dependent mappings include
listwise judging, within-group normalization, abstention/tie handling, and aggregation of per-step scores into
trajectory-level targets. (When deterministic checkers exist, $\mathrm{Eval}$ can be instantiated as a verifier.)

At the highest level, we recover the training signal notation from \S\ref{sec:notations} via
\begin{equation}
S(\tau^{(i)};\mathcal{T}(x)) \;=\; \mathrm{Score}\!\left(\phi_i\right),
\label{eq:f-S}
\end{equation}
with the understanding that $S$ may correspond to $\tilde r_i$, $\tilde A_i$, $\ell_i$, or a weighted objective defined by
$(w_i,\cdot)$. This formulation cleanly distinguishes \textbf{Filter} (measurement and signal construction) from
\textbf{Control} (budgeted decision rules that consume these signals).

\input{figures/filter_tables}

\subsection{Structural Validity (Pre-Semantic Gates) }
\label{sec:f_structural_validity}

Pre-semantic gating filters out rollouts that cannot be meaningfully scored \emph{even if} their underlying intent is correct.
In practice, verifiers are brittle to malformed outputs: a correct solution expressed with the wrong schema, an invalid tool call,
or an unparsable final answer can be misclassified as incorrect and inject systematic noise into training. Structural validity gates
therefore enforce constraints such as: required delimiters (e.g., separating \emph{analysis} from \emph{final}), single-answer policies,
JSON/schema validity for tool outputs, argument typing for function calls, or syntactic constraints on programs and queries.

Recent verifier toolkits emphasize that \emph{normalization and parsing} are first-order components of correctness.
For example, xVerify~\citep{chen2025xverify} focuses on efficient and standardized answer verification for reasoning evaluations,
while Math-Verify provides practical utilities for robust mathematical parsing and comparison~\citep{mathverify}.
Although these tools are often presented for evaluation, in rollout-based RL they play a training-critical role:
they reduce false negatives caused by superficial formatting mismatches and make the mapping from rollouts to scores more stable.

Structural gating is especially salient for execution-centric domains. In program synthesis and code generation,
rollouts must at minimum compile/execute before unit-test feedback can be computed, as in RL from unit test feedback~\citep{liu2023rltf} and execution-guided code RL~\citep{le2022coderl}. Similarly, for Text-to-SQL, a candidate query must be executable
against the database before any semantic comparison is possible; benchmarks such as BIRD emphasize execution-grounded evaluation for realistic text-to-SQL~\citep{li2023llmservedatabaseinterface}, motivating pipelines that treat executability as a first-stage filter
before applying execution-based rewards (e.g.,~\citep{yao2026arctictext2sqlr1simplerewardsstrong,ali2025stateoftheartsqlreasoningmodel}). In short, pre-semantic gates define
the \emph{eligible rollout set} on which meaningful verification and learning can occur.

\subsection{Correctness Verification (Hard Semantic Gates)}
\label{sec:f-verification}

Hard semantic gates assign rollouts task-grounded correctness signals---typically pass/fail or a bounded scalar score---using
deterministic procedures whenever possible. This paradigm is central to verifiable reinforcement learning (RLVR): rather than training a
learned reward model, the system relies on checkers that are (relatively) objective, cheap, and difficult to exploit.

\paragraph{Execution-Based Verification.}
In code tasks, a natural verifier executes the generated program and computes success from unit tests.
RLTF uses unit-test outcomes as feedback signals for learning~\citep{liu2023rltf}, and CodeRL demonstrates that execution-based rewards
can effectively guide policy improvement in code generation~\citep{le2022coderl}. The rollout strategy and verifier are tightly coupled: one typically samples multiple candidate programs, discards non-executing ones, and reinforces programs that pass more tests.

Text-to-SQL follows an analogous pattern. Since SQL has a well-defined execution semantics, correctness can be assessed by running the query
against the target database and comparing results or using execution success as a proxy. BIRD highlights the importance and difficulty of
execution-grounded SQL evaluation in practice~\citep{li2023llmservedatabaseinterface}. Recent RL-style SQL systems explicitly leverage these hard verifiers,
including Arctic-Text2SQL-R1, which shows strong SQL reasoning from simple, execution-grounded rewards~\citep{yao2026arctictext2sqlr1simplerewardsstrong}, as well as other
execution-based reinforcement approaches~\citep{ali2025stateoftheartsqlreasoningmodel}. These systems treat verification not as a post-hoc metric,
but as the mechanism that determines which rollouts are considered correct enough to drive updates.
ExeSQL strengthens this paradigm for realistic deployments by targeting SQL dialect differences and using execution to filter candidates and construct preference-training signals for Text-to-SQL~\citep{zhang2025exesql}.

\paragraph{Exact-Answer Verification for Math.}
For many math datasets, correctness can be defined by exact-answer checks with careful normalization and parsing, enabling scalable
pass/fail reward signals. MATH provides a large set of competition-style problems~\citep{hendrycks2021measuring}, and OlympiadBench further tests
high-difficulty mathematical reasoning~\citep{he2024olympiadbench}. In these settings, the verification stage often reduces to robustly extracting
the final answer and checking equivalence, which again highlights the importance of parsing infrastructure such as xVerify~\citep{chen2025xverify} and Math-Verify~\citep{mathverify}
for avoiding spurious failures.

\paragraph{Multimodal Settings and Making Verification Possible.}
Compared to text-only domains, multimodal reasoning often lacks universal deterministic checkers. Recent work, therefore, either (i) designs
objectives with more verifiable structure or (ii) constructs synthetic pipelines that retain programmatic checkability.
Vision-R1 and VLM-R1 study R1-style reinforcement learning with verifiable rewards for multimodal models; VLM-R1 in particular emphasizes training stability and generalization under visual complexity~\citep{huang2025visionr1incentivizingreasoningcapability,shen2025vlmr1stablegeneralizabler1style}. For video spatial reasoning, SpaceR similarly relies on rule-based verifiable rewards to provide reliable training signals~\citep{ouyang2025spacer}. A complementary direction is to expand the space of tasks with reliable filters through
verifiable synthesis: SynthRL scales visual reasoning by generating data with built-in verifiability~\citep{wu2025synthrl}. 
GRACE \citep{sun2025grace} introduces a verification reward based on contrastive learning.
Across these works, the hard-gate verifier (or its proxy) determines which parts of the rollout distribution are learnable under RL.

\subsection{Process Quality Scoring (Dense or Step-Level Signals) }
\label{sec:f-process}

Beyond binary acceptance or rejection, \emph{process quality scoring} assigns dense signals along a rollout per step, segment, or node, enabling early pruning, partial credit, and better credit assignment. A central distinction is between outcome reward models ORMs that score only the terminal answer, and process reward models PRMs that evaluate intermediate reasoning steps~\citep{liu2025enhancing}. Early empirical evidence on GSM8K shows why this matters---optimizing only for final answer correctness can leave substantial trace error, motivating supervision or learned rewards that track step correctness~\citep{uesato2022solving}. PRM style supervision can be instantiated directly via human step labels, as in PRM800K, where annotators mark individual steps, for example, positive, neutral, negative, to train step-level reward or prediction heads \citep{lightman2023let}. Complementarily, automated step verifiers can provide process feedback without human annotations, for example, by verifying intermediate reasoning steps and using them as reinforcement signals \citep{wang2023math}. In rollout systems, these step-level scores can be aggregated, for example, minimum or first failure, discounted sum, or calibrated averages to decide whether to continue expanding a branch and to shape advantages for downstream optimization~\citep{liu2025enhancing}.

A key challenge is \emph{evaluating} whether dense scores truly reflect reasoning quality rather than correlating with surface heuristics. \citet{zheng2024processbench} address this with \textsc{ProcessBench}, which measures a model’s ability to identify the \emph{earliest erroneous step} in mathematical solutions annotated by experts. This reframes PRM assessment from indirect end task improvements, for example, best-of-$K$ reranking, to direct error localization and calibration, and it highlights a recurring pathology: on harder problems, models can reach correct final answers while still containing incorrect intermediate steps, making terminal-only rewards an unreliable proxy for process fidelity~\citep{zheng2024processbench,uesato2022solving}. Practically, these findings justify deploying process scorers as filters inside the rollout, not just as final rerankers, for example, to cut off low-quality branches early, allocate more budget to uncertain steps, or route suspect segments to stronger critics~\citep{liu2025enhancing}.

Two complementary lines of work further sharpen the evaluation story. PRMBench introduces a fine-grained benchmark designed specifically to probe whether PRMs assign correct step-level signals under challenging failure modes~\citep{song2025prmbench}. Separately, The Lessons of Developing Process Reward Models in Mathematical Reasoning~\citep{zhang2025prmlessons} reports practical pitfalls in PRM development and evaluation, emphasizing that seemingly reasonable estimation or selection procedures can yield misleading conclusions if the benchmark or protocol is not carefully controlled.

Rollouts can be filtered using \emph{offline} structure that scores reasoning traces without requiring a fresh on-policy sample at evaluation time.
OCEAN models chain-of-thought as an MDP, uses knowledge-graph exploration to simulate preferences over reasoning paths, and constructs inverse-propensity-style estimates that turn logged CoT into alignment signals~\citep{wuocean}.
Analysis-oriented filters also treat CoT as structured dynamics: state-aware clustering of steps yields interpretable diagnostics for consistency and transition patterns~\citep{yu2025explainable}, while causal-intervention views help strip spurious correlations between pretraining-induced bias and surfaced reasoning when mapping traces into training targets~\citep{wu2024decot}.

Dense scoring can also be induced through \emph{reference-based} verification systems that convert task-specific checkers or verifiers into step- or node-level feedback. 
\citet{yan2025verifybench} benchmark such reference-based reward systems and stress-test robustness, while \citet{chen2025xverify} provides a verifier designed to improve answer extraction or equivalence checking, crucial whenever correctness depends on parsing or semantic normalization rather than exact match. 
On the optimization side, recent RL methods explicitly operationalize step-level rewards. 
R1-VL~\citep{zhang2025r1} introduces step-wise, verifier-driven reward shaping for multimodal reasoning, converting terminal evaluation into denser step-level signals to mitigate sparse feedback. 
TreeRL~\citep{hou2025treerlllmreinforcementlearning} uses on-policy tree search with intermediate supervision to propagate evaluative signals to internal nodes. 
Rule-based verifiers (e.g., Math Verify) and tree-sampling credit-assignment variants (e.g., TreeRPO~\citep{yang2025treerpo}) convert sparse task feedback into dense rollout control signals.

\subsection{Comparative Assessment (Relative Ranking Signals) }
\label{sec:f-comparative}

Comparative assessment evaluates candidate rollouts through relative judgments rather than absolute scores. Given multiple responses for the same prompt, an evaluator outputs preferences among candidates, which is useful when correctness is ambiguous, when goals are preference aligned, or when selection must be robust across diverse samples. This mechanism naturally acts as a filtering signal that can later be converted into supervision for training or used directly for pruning and selection \citep{zheng2023judging}.

A common approach uses an LLM as a judge to compare candidates, but such judgments can be noisy and systematically biased. MT Bench and Chatbot Arena document sensitivity to answer ordering, preference for longer outputs, and weaknesses on reasoning and mathematics \citep{zheng2023judging}. Judging the Judges analyzes both pairwise and listwise comparison settings and recommends protocols that improve reliability, including repeated judgments, order permutations, and explicit tie options \citep{shi-etal-2025-judging}. Complementary work studies how feedback protocols (e.g., pairwise vs.\ pointwise) interact with bias in LLM-based evaluation \citep{tripathi2025pairwise}, with broader discussion summarized in recent surveys \citep{gu2024llmasajudge}. Incorporating ties and abstention is also important when differences are not decisive, since models may otherwise produce overconfident rankings, and abstention behavior can degrade under reasoning focused fine-tuning~\citep{shi-etal-2025-judging,kirichenko2025abstentionbenchreasoningllmsfail}.
Recent work also attempts to \emph{improve} judge reliability directly at inference time by leveraging the judge’s \emph{distribution} over judgment tokens rather than relying on a single greedy decision~\citep{wang2025judgmentdistribution}. For broader context and taxonomy, see recent surveys of LLM-as-a-judge~\citep{li2025generation}.

Comparative assessment can be strengthened by attaching critiques that justify preferences and help identify failure modes. Critic CoT~\citep{criticcot} demonstrates the value of using critique-based filtering before aggregation, while Critic RM~\citep{criticrm} uses critique generation and consistency filtering to improve downstream reward modeling.

\subsection{Learning-Value Signals (Informativeness, Not Correctness) } \label{sec:f-learningvalue}

Beyond assigning scores based on semantic correctness, a distinct class of filtering signals assesses the \emph{informativeness} or \emph{learning value} of a rollout or prompt. These signals aim to maximize training progress by identifying data points where the model’s current policy is uncertain, where reward signals are most variable, or where additional exploration is likely to yield novel, generalizable behaviors. This paradigm is crucial for overcoming the gradient-vanishing problem common in group-based RL such as GRPO, where uniform rewards (all correct or all incorrect) within a group lead to zero advantage and thus no learning signal.

A primary strategy is to use the \emph{variance of group rewards} as a real-time, adaptive proxy for sample difficulty and learning potential. The core insight is that prompts yielding a mix of correct and incorrect rollouts, feature high reward variance, lie near the model's current capability boundary and provide the strongest contrastive gradients. VCRL~\citep{jiang2025vcrl} formalizes this by constructing a curriculum that dynamically prioritizes prompts with high normalized reward variance, using a replay buffer to maintain a pool of high-value samples. Similarly, DAPO~\citep{yu2025dapo} employs dynamic sampling, discarding prompts with uniform rewards and oversampling until a batch with informative variance is assembled, ensuring every training step receives non-zero gradients. Extending this to a non-stationary bandit formulation, VADE~\citep{hu2025vade} performs online, sample-level difficulty estimation using a Beta distribution posterior. It selects prompts via Thompson sampling to maximize an information gain objective, effectively balancing exploration and exploitation without extra rollout costs.

A complementary approach quantifies uncertainty through the \emph{semantic entropy} of generated responses, measuring the diversity of meanings within a rollout group. SEED-GRPO~\citep{chen2025seedgrposemanticentropyenhanced} uses this entropy to modulate advantage magnitudes: prompts eliciting semantically consistent responses that have low entropy and high confidence undergo standard updates, while those with diverse, contradictory outputs that have high entropy and high uncertainty receive attenuated updates, implementing a dynamic, uncertainty-aware learning rate. XRPO~\citep{bamba2025xrpo} integrates multiple informative signals. Its hierarchical rollout planner allocates compute based on a priority score combining expected statistical uncertainty reduction and an exploration bonus, while its novelty-guided advantage sharpening rewards correct but low-likelihood responses, pushing the policy boundary outward.

Efficiency-oriented methods focus on \emph{reusing or down-sampling} rollouts to preserve information gain while reducing computation. AR3PO~\citep{zhang2025improvingsamplingefficiencyrlvr} introduces adaptive rollout that allocates more responses to hard prompts and response reuse that leverages past correct responses from a buffer, significantly lowering the average number of rollouts needed per training step. PODS~\citep{xu2025rolloutsusefuldownsamplingrollouts} addresses the system-level asymmetry between parallelizable inference and memory-intensive policy updates by generating many rollouts but training only on an informative subset. Its max-variance down-sampling selects a group of rollouts that maximize reward variance within a group, which, for binary rewards, simplifies to choosing the half group of best and half group of worst, preserving strong contrastive signals at a fraction of the update cost.

Relatedly,~\citet{nguyen2026adaptiverolloutallocationonline} studies online RLVR with a policy that dynamically allocates rollout budget based on observed learning signals, making compute allocation itself a learned, feedback-driven component of the training loop.

GRPO-LEAD~\citep{zhang2025grpo} incorporates a difficulty-aware advantage reweighting mechanism. It estimates prompt difficulty via the empirical correctness ratio of its rollouts and applies a logistic weighting function to amplify advantages for challenging problems, ensuring the model focuses on refining its performance on harder tasks rather than over-optimizing on easy ones. Collectively, these learning-value signals transform the filtering stage from a passive correctness check into an active engine for directing exploration, stabilizing gradients, and maximizing the informational yield of each unit of rollout computation.

\subsection{Training-Signal Construction (Scores, Labels, Advantages, or Weights) }
\label{sec:f-supervision}

Even when a hard verifier provides a clean score $r_i^{\mathrm{raw}}$, training typically does not optimize directly on $r_i^{\mathrm{raw}}$.
Instead, systems transform verification outcomes into learning targets that improve optimization stability, reduce variance, and
encourage useful exploration. This mapping is particularly important in rollout-heavy settings, where most samples may be incorrect
and naïve credit assignment can lead to collapse.

\paragraph{Group-Relative Advantages from Multiple Rollouts.}
A widely used strategy is to sample multiple rollouts per prompt and compute \emph{within-group} baselines to normalize difficulty.
For example, one can define an advantage by subtracting the mean score among valid rollouts:
\begin{align}
\tilde{A}_i
\;=\;
m_i\Bigg(
r_i^{\mathrm{raw}}
-
\frac{\sum_{j=1}^{K} m_j\, r_j^{\mathrm{raw}}}{\sum_{j=1}^{K} m_j + \epsilon}
\Bigg),
\end{align}
where $\epsilon>0$ is a small constant for numerical stability.

So updates depend on \emph{relative} performance among candidate trajectories for the same $x$.
Analyses of GRPO objectives clarify how such group-based losses behave and how rollout sets influence the effective optimization signal~\citep{mroueh2025reinforcementlearningverifiablerewards}. When exploration diminishes (e.g., low-entropy rollout sets), the mapping from $r_i^{\mathrm{raw}}$ to $\tilde{A}_i$ becomes even
more consequential; GRPO-LEAD~\citep{zhang2025grpo} studies advantage reweighting strategies aimed at maintaining learning progress under such conditions.

\paragraph{Filtering, Shaping, and Verbosity Control.}
Beyond baselines, the mapping stage can incorporate additional shaping terms and selective filtering.
GFPO~\citep{shrivastava2025samplethinklessgroup} exemplifies group filtering paired with concision-aware signal design, discouraging pathological verbosity while preserving
correctness-driven learning. XRPO~\citep{bamba2025xrpo} provides another perspective on constructing training signals that balance targeted
exploration and exploitation using rollout-derived supervision. These works reflect a broader trend: rollout strategy is
not only about how to sample, but also how to \emph{value} and \emph{weight} samples once obtained.

\paragraph{Conservative Optimization under Aggressive Filters.}
Hard gating can produce sparse effective batches (few surviving rollouts), which increases gradient variance and can destabilize training.
Trust-region or conservative-update principles help mitigate this by limiting how far the policy moves per iteration under noisy signals.
TROLL represents this direction, emphasizing stability of policy updates in the presence of filtered or high-variance rollout supervision~\citep{becker2025trolltrustregionsimprove}.

\paragraph{When Hard Verifiers Are Missing: Constructing Surrogate Signals.}
Many real tasks lack deterministic correctness checks, so recent work explores how to retain the benefits of rollout filtering while
broadening applicability. \citet{su2025crossingrewardbridgeexpanding} argues for expanding verifiable-style training across more diverse domains, and Generalizing Verifiable Instruction Following studies how to extend verifiable objectives beyond narrowly
rule-based settings~\citep{pyatkin2025generalizing}. Other efforts explicitly address learning without reliable external verifiers, including Reinforcing
General Reasoning without Verifiers~\citep{zhou2025reinforcing}, RLPR~\citep{yu2025rlpr}, and Zero Reinforcement Learning~\citep{zeng2025zero}.
In these settings, the \emph{filter} often becomes a hybrid of weak checks, self-consistency signals, or structured constraints, and the
signal-construction stage carries more responsibility for stabilizing learning and preventing reward hacking.

%% file: figures/filter_tables.tex
\begin{table}[ht]
\centering
\small
\setlength{\tabcolsep}{4pt}
\renewcommand{\arraystretch}{1.15}
\caption{\textbf{Filter as a signal pipeline (Gate $\rightarrow$ Eval $\rightarrow$ Map).} This table instantiates Eqs.~(\ref{eq:f-gate})--(\ref{eq:f-map}) with common rollout filters and the optimizer-facing artifacts they produce.}
\label{tab:filter_gate_eval_map}
\begin{tabularx}{\linewidth}{@{} p{0.05\linewidth} p{0.12\linewidth} p{0.40\linewidth} Y @{}}
\toprule
\rowcolor{gfcrHead}
\textbf{Stage} & \textbf{Output} & \textbf{Typical mechanisms} & \textbf{Representative works} \\
\midrule
\rowcolor{gfcrF!70}
\textbf{Gate} & $m_i \in \{0,1\}$ &
Pre-semantic validity checks: parseability/normalization, executability, safety constraints; can act on rollouts or prefixes. &
\citep{chen2025xverify,mathverify,liu2023rltf,le2022coderl,li2023llmservedatabaseinterface}. \\
\hline
\rowcolor{gfcrG!65}
\textbf{Eval} & $r_i^{\mathrm{raw}} \in \mathbb{R}$ &
Hard semantic verification (unit tests, execution, exact-answer equivalence), process/step scoring (PRMs/step verifiers), or comparative ranking (LLM judge). &
\citep{liu2023rltf,le2022coderl,yao2026arctictext2sqlr1simplerewardsstrong,ali2025stateoftheartsqlreasoningmodel,zhang2025exesql,hendrycks2021measuring,he2024olympiadbench,lightman2023let,zheng2024processbench,song2025prmbench,zhang2025prmlessons,zheng2023judging,shi-etal-2025-judging}. \\
\hline
\rowcolor{gfcrC!55}
\textbf{Map} & $(w_i,\tilde r_i,\tilde A_i,\ell_i)$ &
Convert raw signals into optimizer-facing supervision: masks/weights, shaped rewards, group-relative advantages, pairwise/listwise labels; often group-dependent. &
\citep{mroueh2025reinforcementlearningverifiablerewards,zhang2025grpo,shrivastava2025samplethinklessgroup,becker2025trolltrustregionsimprove}. \\
\bottomrule
\end{tabularx}
\end{table}

%% file: latex/4.3_control.tex
\section{Control: Compute Allocation, Decision Rules, and On/Off-Policy Knobs} \label{sec:c}

\begin{figure}[ht]
    \centering
    \includegraphics[width=1.0\linewidth]{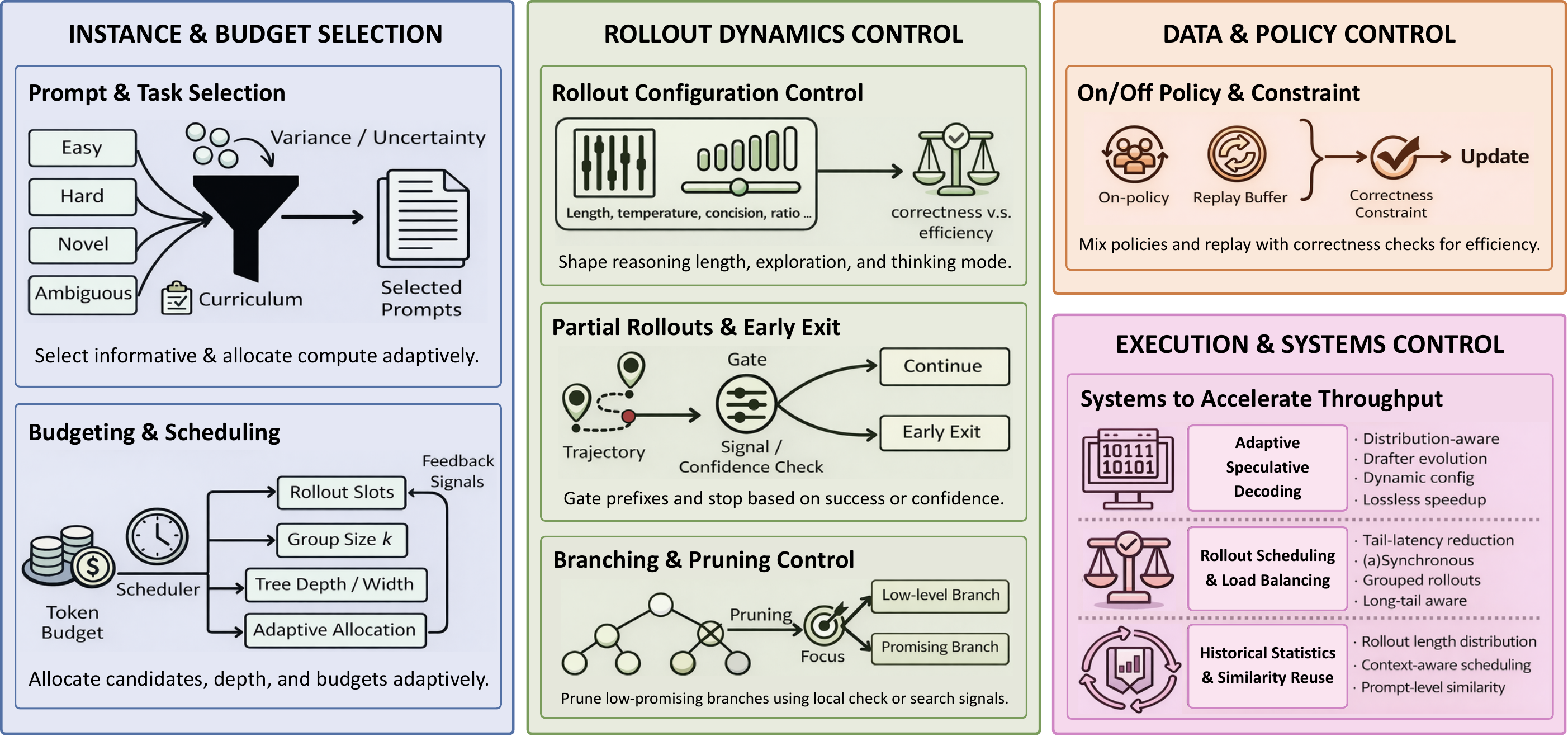}
    \caption{\textbf{Control module design space.} The control layer of rollout pipelines can be organized into four orthogonal components: \emph{(left)} Instance \& Budget Selection (prompt prioritization by difficulty/novelty/uncertainty; adaptive allocation of rollout slots, group size, depth/width, and token budgets); \emph{(middle)} Rollout Dynamics Control (decoding configuration, partial rollouts with early exit, and branching–pruning policies for quality–efficiency trade-offs); \emph{(right-top)} Data \& Policy Control (on-/off-policy mixing, replay, and correctness-aware filtering); and \emph{(right-bottom)} Execution \& Systems Control (scheduling, speculative decoding, load balancing, and statistics reuse). Together, these mechanisms manage compute allocation and execution under fixed budgets, defining the trade-off between accuracy, efficiency, and cost independently of the underlying optimizer.}
    \label{fig:framework-control}
\end{figure}

\subsection{Problem Formulation: Control (C) --- Compute Allocation and Decision Rules}
\label{sec:c-formulation}

Using the global notation in \S\ref{sec:notations}, the \textbf{Control} module governs \emph{where to spend rollout compute
and what to do next} under budgets: which prompts/tasks to roll out on, how many candidates to generate, when to stop or
continue, when to branch or prune, and how to set rollout configuration (e.g., max length, temperature, concision). Control
turns intermediate Filter signals and cost accounting into allocation decisions, thereby shaping the realized rollout-group
distribution $q_{\theta,\mathrm{GFCR}}(\mathcal{T}\mid x,\mathcal{B})$.

Control can be viewed as a budgeted sequential decision process operating over an evolving set of partial rollouts. For a
prompt $x$ and an active frontier of prefixes $\{\tau_{\le t}^{(i)}\}$, let
$\phi_{i,t} := \phi(\tau_{\le t}^{(i)};\mathcal{T}(x))$ denote the associated Filter signal(s) available at time $t$ (e.g., step scores,
validity flags, verifier outcomes). Control observes the frontier, these signals, accumulated cost, and remaining budget
$b$ (initialized as $B_x$), and selects an action
\begin{equation}
a \sim C_\psi(\cdot \mid x, \{\tau_{\le t}^{(i)}\}, \{\phi_{i,t}\}, b),
\label{eq:c-policy}
\end{equation}
where $a$ may encode: (i) prompt/task selection across instances, (ii) adaptive sampling decisions (e.g., choosing $K$,
depth/width, or selective rollout), (iii) continuation/stop rules (early exit, partial rollouts, selective continuation),
(iv) branch/prune control in tree/agentic rollouts (which prefixes to expand and which subtrees to discard), and (v) rollout
configuration control (length/temperature/concision and sampling/selection rules that affect the mix of positive vs.\
informative-negative trajectories). These actions update the frontier, determine additional token/tool expenditure, and
ultimately determine the terminal rollout group $\mathcal{T}(x)$ passed to downstream learning-signal construction.

At the pipeline level, Control aims to maximize learning utility per unit compute by shaping which rollouts are completed
and which are abandoned:
\begin{equation}
\max_{\psi}\;\; \mathbb{E}_{x \sim \mathcal{D}}\;
\mathbb{E}_{\mathcal{T} \sim q_{\theta,\mathrm{GFCR}}(\cdot \mid x,\mathcal{B};\psi)}\!\left[U(\mathcal{T})\right]
\quad \text{s.t.}\quad
\mathbb{E}\!\left[\sum_{\tau \in \mathcal{T}} c(\tau)\right] \le B_x,
\label{eq:c-objective}
\end{equation}
where $U(\mathcal{T})$ summarizes the downstream training value induced by Filter (e.g., usable-sample mass, expected signal
strength, or other signal-quality surrogates), and the constraint enforces per-prompt compute limits (with an analogous
global constraint $B$). Finally, Control includes on/off-policy knobs that govern how training mixes fresh on-policy rollouts
with reused/off-policy data (often mediated by Replay), as well as systems-level scheduling/token accounting decisions that
treat throughput and long-tailed rollout lengths as first-class control problems. Table~\ref{tab:c_control_map} summarizes
the method landscape covered across this entire section.

\input{tables/control}

\subsection{Prompt and Task Selection (What Do You Roll Out On?)\label{sec:c-task}}
An important but sometimes overlooked design choice in reinforcement learning based post training is deciding which prompts should be used to generate rollouts. Early work often relied on uniform sampling from the training distribution. However, several recent studies show that many prompts contribute little useful training signal, which leads to wasted computation. GRESO~\citep{zheng2025actpaysefficientreinforcement} targets this inefficiency by observing that when all samples in a rollout group receive identical rewards, the within-group reward variance becomes zero and GRPO-style advantages collapse, yielding near-zero policy-gradient signal. Based on training dynamics, GRESO~\citep{zheng2025actpaysefficientreinforcement} learns to predict such \emph{zero-variance} prompts and probabilistically skips them before rollout while retaining exploration. VCRL~\citep{jiang2025vcrl} takes a complementary view by treating group reward variance as a proxy for sample difficulty: prompts that are too easy or too hard tend to have low variance, while intermediate-difficulty prompts produce higher variance and stronger learning signal, inducing an implicit curriculum as the policy improves.\par
Other work has explored uncertainty driven prompt selection using probabilistic modeling. VADE~\citep{hu2025vade} estimates per-prompt correctness online (via Beta posteriors) and applies Thompson sampling to favor prompts expected to be informative while still exploring. In contrast, SEED-GRPO~\citep{chen2025seedgrposemanticentropyenhanced} does not explicitly decide which prompts to sample; instead, it measures semantic entropy (meaning diversity) across multiple answers to a prompt and uses this uncertainty signal to modulate the magnitude of policy updates, applying more conservative updates on high-uncertainty prompts.\par
SEC~\citep{chen2025selfevolvingcurriculumllmreasoning} moves further away from individual prompt level selection and instead learns a distribution over prompt categories. It formulates curriculum selection as a non stationary multi armed bandit problem, where each category such as difficulty level or task type is treated as an arm; the absolute advantage acts as the reward signal and the curriculum policy is updated (e.g., via TD(0)) to maximize immediate learning gain. Taken together, these approaches reflect a broader shift toward viewing rollout allocation as an adaptive resource management problem rather than a fixed data sampling procedure.

\subsection{Budgeting and Scheduling (How Much Do You Roll Out?)\label{sec:c-budget}}
One recurring design choice across recent reinforcement learning approaches for reasoning is how rollout compute is distributed during training. Early GRPO-style methods (e.g., \cite{Guo_2025}) often used a fixed number of sampled trajectories per prompt to form group-relative advantages. Several works instead scale rollout compute directly by increasing either the number (breadth) or the repeated optimization depth on a prompt. For example, DeepSeekMath \citep{shao2024deepseekmath} uses GRPO-style group sampling (multiple candidates per question) with within-group reward normalization, directly scaling compute with group size. Similarly, 1-shot RLVR \citep{wang2025reinforcementlearningreasoninglarge} repeatedly samples rollouts and updates on a single training example over many RL steps, suggesting that large per-prompt rollout depth can partially substitute for dataset diversity.\par
More recent methods argue that rollout budgets should be allocated adaptively rather than uniformly across prompts. Variance-aware scheduling approaches such as VIP \citep{nguyen2026adaptiverolloutallocationonline} allocate rollout compute to minimize expected policy-gradient variance using predicted per-prompt success probabilities. MMR1 \citep{leng2025mmr1enhancingmultimodalreasoning} similarly biases sampling toward prompts with higher outcome variance and trajectory diversity to mitigate gradient vanishing. XRPO \citep{bamba2025xrpo} introduces a rollout allocator that prioritizes prompts with higher potential for uncertainty reduction, and further improves exploration/exploitation via in-context seeding on zero-reward prompts and novelty-aware advantage sharpening. A complementary line of work focuses on reducing rollout cost through adaptive rollout and reuse: AR3PO \citep{zhang2025improvingsamplingefficiencyrlvr} allocates more responses to difficult prompts while saving computation on easier ones, and reuses previously generated correct responses to provide additional training signal.

\subsection{Rollout Configuration Control (Length, Temperature, Concision, Positive or Negative Ratio)  } \label{sec:c-config}

A key control knob in reasoning RL is how rollouts are parameterized and selected under a compute
budget. Recent work shows that longer Chain-of-Thought traces are not uniformly beneficial:
while they may help with hard problems, they often induce redundant \emph{overthinking} that inflates the token
usage without improving correctness. Accordingly, configuration control increasingly treats reasoning length and thinking mode as decision variables, and shapes the training experience via sampling, filtering, and curricula to trade train-time compute for test-time efficiency.

Several methods directly optimize for shorter-but-correct trajectories. ShorterBetter \citep{yi2025shorterbetter} defines the Sample Optimal Length (SOL)---the shortest correct response among multiple generations---and uses it as a dynamic reward signal to learn an instance-adaptive optimal CoT length without manual supervision. Addressing the limitation of naive length penalties, DECS \citep{jiang2025overthinkingreductiondecoupledrewards} identifies a mismatch between trajectory-level rewards and token-level optimization, and introduces decoupled token-level rewards together with curriculum batch scheduling to penalize redundant tokens while preserving essential exploration. WS-GRPO~\citep{mundada2026wsgrpoweaklysupervisedgrouprelativepolicy} takes a weakly supervised approach, training a preference model from correct versus incorrect rollouts and using it to compare consecutive prefixes, yielding correctness-aware length control without explicit penalty tuning.

Another line of work controls \emph{whether} to think. AdaptThink \citep{zhang2025adaptthinkreasoningmodelslearn} observes that a direct-answer mode (NoThinking) can outperform long reasoning on simpler queries, and proposes an RL algorithm that learns to select between thinking modes based on problem difficulty while maintaining overall performance. Similarly, Large Hybrid-Reasoning Models employ a two-stage pipeline (cold-start fine-tuning followed by online RL) to learn hybrid thinking decisions and evaluate this capability with a dedicated Hybrid Accuracy metric \citep{jiang2025think}.
CoRL \citep{jin2025controlling} introduces a reinforcement learning framework that optimizes the performance–cost trade-off when reasoning with an external LLM.

Finally, configuration control can be realized through train-time sampling and selection rules that
shape which trajectories drive updates. GFPO \citep{shrivastava2025samplethinklessgroup} mitigates RLVR-induced length inflation by sampling larger groups per problem and filtering training responses using length and token-efficiency (reward per token), demonstrating a direct trade-off where increased training-time compute yields reduced test-time compute. Complementarily, Train Long, Think Short applies a length-control curriculum under GRPO, starting with generous token budgets and progressively tightening them, with rewards balancing correctness, length efficiency, and formatting adherence, consistently outperforming fixed-budget baselines at the same final budget \citep{hammoud2025train}.

\subsection{Partial Rollouts, Early Exit, and Continuation Rules}\label{sec:c-stop}

Control can reduce rollout cost by (i) learning when to stop reasoning within a trajectory, (ii) pausing/continuing partial generations to mitigate long-tail stragglers, or (iii) down-selecting which rollouts participate in the (communication-heavy) policy update. S-GRPO~\citep{dai2025s} targets early exit by sampling a single reasoning path and training the model to exit at earlier positions via serial-group decaying rewards, encouraging concise thoughts and earlier termination. APRIL~\citep{zhou2025aprilactivepartialrollouts} is a systems-oriented partial-rollout scheme that over-provisions rollout requests, terminates a batch once enough responses are completed, and recycles unfinished generations for continuation in future steps, improving GPU utilization under long-tail response lengths. PODS~\citep{xu2025rolloutsusefuldownsamplingrollouts} reduces update cost by selecting a strategically chosen subset of rollouts (e.g., max-variance down-sampling that maximizes reward diversity) to train on, decoupling cheap rollout generation from expensive multi-device optimization. WS-GRPO~\citep{mundada2026wsgrpoweaklysupervisedgrouprelativepolicy} derives prefix-level continue/stop signals from a preference model trained on outcome correctness, converting preference margins between consecutive prefixes into dense pseudo-rewards that reduce redundant continuation without requiring a global length penalty.

\subsection{Branching and Pruning Control (Trees and Agents)  } \label{sec:c-branch}

Branching and pruning strategies control how compute is allocated across multiple reasoning paths in tree-structured rollouts, deciding which nodes to expand, which paths to prune, and how deep to explore under a fixed budget. Beyond improving exploration, tree topology can also be exploited to provide denser supervision by exposing intermediate states for credit assignment. TreeRL~\citep{hou2025treerlllmreinforcementlearning} directly incorporates on-policy tree search into RL training, strategically branching from high-uncertainty intermediate steps to improve search efficiency and provide intermediate supervision without a separate reward model. TreePO~\citep{li2025treepo} treats sequence generation as a tree search with segment-wise decoding; it uses local uncertainty to warrant additional branches, amortizes compute across shared prefixes, and prunes low-value paths early to reduce KV-cache and sampling cost. DeepSearch~\citep{wu2025deepsearch} integrates Monte Carlo Tree Search (MCTS) directly into RLVR training, embedding structured search into the training loop to improve exploration and enable fine-grained credit assignment. Finally, TreeRPO~\citep{yang2025treerpo} uses tree sampling to estimate expected rewards at different reasoning steps and computes group-relative step-level rewards, producing dense training signals without training a separate step reward model.

\subsection{On- and Off-Policy Controls and Correctness Constraints} \label{sec:c-offpolicy}
Recent work on reinforcement fine-tuning of large language models highlights the importance of balancing on-policy and off-policy rollouts to maintain both training stability and distributional correctness. On-policy rollouts align updates with the current policy but are compute intensive; replay and mix-policy approaches aim to reuse past rollouts to improve data efficiency. RePO~\citep{li2025repo} augments GRPO with a replay buffer and replay strategies to reuse off-policy rollouts, improving sample efficiency while retaining an on-policy component. ReMix~\citep{liang2025squeeze} proposes Reincarnating Mix-policy Proximal Policy Gradient (ReMix), a general approach that enables on-policy RFT methods such as PPO and GRPO to leverage off-policy data for more compute-efficient training.

\subsection{Systems to Accelerate Rollout Throughput } \label{sec:c-systems}

Systems for accelerating rollout throughput treat long rollouts and batched generation as first-class control problems, explicitly optimizing token usage, scheduling, and infrastructure constraints in reinforcement learning pipelines. ReSpec~\citep{chen2025respec} adapts speculative decoding to RL by dynamically tuning decoding configurations, evolving the drafter to avoid staleness (e.g., via distillation), and reward-weighting updates to preserve training stability. DAS~\citep{shao2025beat} targets long-tail rollout lengths with a distribution-aware speculative-decoding framework that leverages historical rollout statistics to accelerate generation without changing model outputs. TLT~\citep{hu2025taming} similarly integrates adaptive speculative decoding into reasoning-RL pipelines to reduce long-tail latency and speed up training losslessly. EARL~\citep{tan2025earl} optimizes agentic RL training under rapidly growing multi-turn contexts using dynamic parallelism selection and layout-aware decentralized data dispatch to improve throughput and reduce long-context failures. Finally, Seer~\citep{qin2025seer} improves synchronous RL rollouts by exploiting prompt-level similarities in output lengths and generation patterns, combining divided rollouts for dynamic load balancing, context-aware scheduling, and adaptive grouped speculative decoding to reduce tail latency and raise throughput.

%% file: tables/control.tex
\begin{table}[!ht]
\centering
\caption{\textbf{Taxonomy of control mechanisms in rollout-based RL for reasoning LLMs.} Rows correspond to control loci in \S\ref{sec:c}; columns specify the controlled decision variable, the signals that drive that decision, representative methods, and the operational objective/trade-off targeted by each family.}
\label{tab:c_control_map}
\scriptsize
\setlength{\tabcolsep}{4pt}
\renewcommand{\arraystretch}{1.10}

\begin{tabularx}{\linewidth}{@{} >{\raggedright\arraybackslash}p{0.17\linewidth} >{\raggedright\arraybackslash}p{0.17\linewidth} >{\raggedright\arraybackslash}p{0.18\linewidth} >{\raggedright\arraybackslash}p{0.22\linewidth} X @{}}
\toprule
\rowcolor{gfcrHead}
\textbf{Control slice} & \textbf{Decision variable} & \textbf{Primary signals} & \textbf{Representative methods} & \textbf{Operational objective and trade-off} \\
\midrule

\rowcolor{gfcrG!65}
Prompt/task selection &
Which prompts are rolled out &
Group reward variance, posterior success uncertainty, semantic entropy, category-level learning gain &
GRESO~\citep{zheng2025actpaysefficientreinforcement}; VCRL~\citep{jiang2025vcrl}; VADE~\citep{hu2025vade}; SEED-GRPO~\citep{chen2025seedgrposemanticentropyenhanced}; SEC~\citep{chen2025selfevolvingcurriculumllmreasoning} &
Avoid low-information prompts and concentrate computation where policy updates are expected to be strongest. \\
\hline
\rowcolor{gfcrC!70}
Budgeting and scheduling &
How many rollouts (width/depth) per prompt &
Fixed-$K$ groups vs. variance-aware and uncertainty-aware allocation, response reuse statistics &
DeepSeekMath~\citep{shao2024deepseekmath}; 1-shot RLVR~\citep{wang2025reinforcementlearningreasoninglarge}; VIP~\citep{nguyen2026adaptiverolloutallocationonline}; MMR1~\citep{leng2025mmr1enhancingmultimodalreasoning}; XRPO~\citep{bamba2025xrpo}; AR3PO~\citep{zhang2025improvingsamplingefficiencyrlvr} &
Trade static stability for adaptive compute efficiency, often reallocating budget toward uncertain or harder prompts. \\
\hline
\rowcolor{gfcrF!55}
Rollout configuration control &
Length, think/no-think mode, and token-efficiency of kept responses &
SOL targets, decoupled token rewards, hybrid-mode gating, reward-per-token filters, length curricula &
ShorterBetter~\citep{yi2025shorterbetter}; DECS~\citep{jiang2025overthinkingreductiondecoupledrewards}; AdaptThink~\citep{zhang2025adaptthinkreasoningmodelslearn}; LHRMs~\citep{jiang2025think}; GFPO~\citep{shrivastava2025samplethinklessgroup}; Train Long, Think Short~\citep{hammoud2025train} &
Reduce overthinking and token cost while preserving correctness through reward shaping and curriculum schedules. \\
\hline
\rowcolor{gfcrC!55}
Partial rollouts and early exit &
Stop, continue, resume, or downsample ongoing rollouts &
Prefix-level progress signals, completion counts, diversity/variance among sampled rollouts &
S-GRPO~\citep{dai2025s}; APRIL~\citep{zhou2025aprilactivepartialrollouts}; PODS~\citep{xu2025rolloutsusefuldownsamplingrollouts} &
Cut long-tail rollout waste by terminating unproductive generation early and prioritizing informative trajectories for updates. \\
\hline
\rowcolor{gfcrG!55}
Branching and pruning (tree rollouts) &
Where to branch and which subtrees to prune &
Local uncertainty, search value estimates, and step-level reward propagation &
TreeRL~\citep{hou2025treerlllmreinforcementlearning}; TreePO~\citep{li2025treepo}; DeepSearch~\citep{wu2025deepsearch}; TreeRPO~\citep{yang2025treerpo} &
Increase search efficiency and credit density by allocating compute to promising reasoning branches under fixed budgets. \\
\hline
\rowcolor{gfcrR!70}
On-/off-policy control &
Mix ratio of fresh rollouts vs replayed data &
Replay-buffer provenance/recency and off-policy correction constraints &
RePO~\citep{li2025repo}; ReMix~\citep{liang2025squeeze} &
Improve sample efficiency using replay while constraining off-policy drift and optimization instability. \\
\hline
\rowcolor{gfcrC!40}
Systems-level control &
Throughput, tail latency, and scheduling policy &
Length-distribution statistics, drafter quality, context similarity, and parallelism layout signals &
ReSpec~\citep{chen2025respec}; DAS~\citep{shao2025beat}; TLT~\citep{hu2025taming}; EARL~\citep{tan2025earl}; Seer~\citep{qin2025seer} &
Treat rollout generation as a systems bottleneck and optimize decoding/scheduling to raise effective training throughput. \\
\bottomrule
\end{tabularx}
\end{table}

%% file: latex/4.4_replay.tex
\section{Replay: Retention, Reuse, and Self-Evolution } \label{sec:r}
\subsection{Problem Formulation: Replay (R) --- Retention, Reuse, and Self-Evolution}
\label{sec:r-formulation}

\begin{figure}[ht]
    \centering
    \includegraphics[width=1.\linewidth]{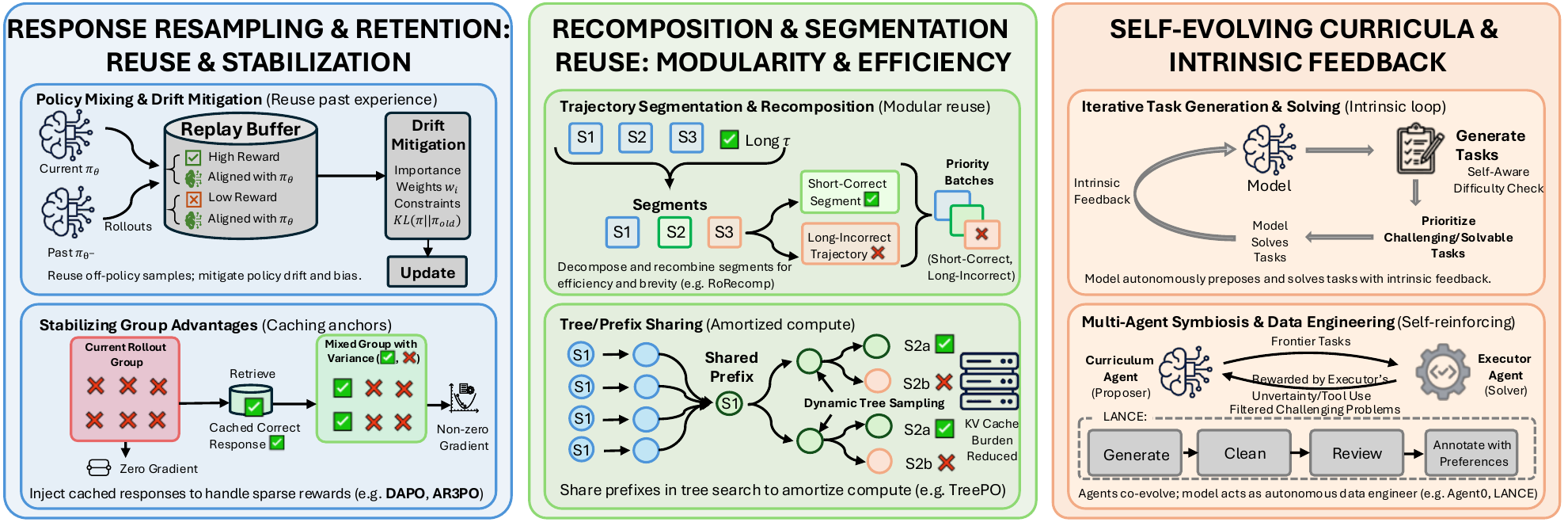}
    \caption{This figure summarizes the Replay layer of rollout pipelines, which governs how past rollouts are retained and reused without updating model parameters. \Cref{sec:r} includes three components: (1) Response Resampling \& Retention, which reuses full trajectories to stabilize learning; (2) Recomposition \& Segmentation Reuse, which reuses segments or shared prefixes for modular efficiency; and (3) Self-Evolving Curricula \& Intrinsic Feedback, which generates new tasks and data through intrinsic feedback. Together, these mechanisms improve data efficiency and training stability.}
    \label{fig:framework-replay}
\end{figure}

Using the global notation in \S\ref{sec:notations}, the \textbf{Replay} module governs what persists across rollouts
\emph{without} updating model parameters. It maintains a persistent replay state $\mathcal{B}$ (buffer/cache) that stores
reusable artifacts derived from past rollouts---cached responses, full trajectories, prefixes $\tau_{\le t}$, verified
segments, or self-generated tasks---so that future rollouts can reuse past computation. Because $\mathcal{B}$ evolves over
time, Replay makes the end-to-end rollout behavior history-dependent: for prompt $x$, the GFCR pipeline induces
\begin{equation}
\mathcal{T} \sim q_{\theta,\mathrm{GFCR}}(\cdot \mid x, \mathcal{B}),
\end{equation}
highlighting that the realized distribution over rollout groups depends not only on $\pi_\theta$ but also on what has been
retained and retrieved as shown in~\Cref{fig:framework-replay}.

Replay can be viewed as inducing an \emph{off-policy} data source. Let $\pi_{\theta^-}$ denote the behavior policy under
which a stored artifact was generated (e.g., a prior checkpoint). Reusing data collected under $\pi_{\theta^-}$ improves
sample efficiency but introduces two core risks under policy drift: (i) \emph{off-policy bias} when learning signals are
computed on mismatched distributions, and (ii) \emph{evaluator drift} when verifiers/judges (or their calibration) change
relative to the current policy outputs. We write each stored entry as
\begin{equation}
e = (\tau, \phi, S, c(\tau), \theta^-, t_{\mathrm{store}}),
\end{equation}
where $\phi$ and $S$ are the Filter signals and derived training signal, $c(\tau)$ is cost, $\theta^-$ is the policy
version (or identifier) that produced $\tau$, and $t_{\mathrm{store}}$ is a timestamp or age.

Formally, after generating a rollout group $\mathcal{T}(x)$, Replay applies a retention rule that decides what to store:
\begin{equation}
\mathcal{B} \leftarrow R_{\mathrm{store}}\!\big(\mathcal{B}, \mathcal{T}(x), \{\phi(\tau^{(i)};\mathcal{T}(x))\}_{i=1}^{K}\big),
\label{eq:r-store}
\end{equation}
and a corresponding policy for eviction, prioritization, or refresh. We denote by $\rho(e \mid x')$ a retrieval score
(or priority) used to select entries relevant to a new prompt $x'$, which may depend on similarity, verified correctness,
diversity, cost, and freshness:
\begin{equation}
\mathcal{Z}(x') \sim R_{\mathrm{retrieve}}(\mathcal{B}\mid x') \quad \text{with}\quad
R_{\mathrm{retrieve}} \ \text{favoring high } \rho(e \mid x').
\label{eq:r-retrieve}
\end{equation}
Retrieved artifacts $\mathcal{Z}(x')$ can condition \textbf{Generate} (e.g., cached candidates, exemplars, verified
sub-traces), influence \textbf{Control} (e.g., warm-starting the frontier or allocating fewer samples when strong cached
solutions exist), and stabilize \textbf{Filter} (e.g., by maintaining reward variance in group-based objectives).

This abstraction unifies three reuse granularities surveyed in this section. First, \emph{response resampling and retention}
treats full rollouts as reusable units, enabling replay buffers, policy mixing, and mechanisms that stabilize group-based
objectives under policy collapse (e.g., injecting cached correct responses to prevent advantages from vanishing). Second,
\emph{recomposition} treats trajectories as compositional, storing and recombining verified segments or prefixes
$\tau_{\le t}$ to amortize shared computation and target efficiency goals such as reduced verbosity or fewer tool calls.
Third, \emph{self-evolution} goes beyond the reuse of existing data: rollouts generate new tasks, solutions, or agents that feed
back into training, forming self-evolving curricula that effectively expand the task distribution.

\subsection{Response Resampling and Retention}
\label{sec:r-buffer}

Empirical study has demonstrated the effectiveness of replay strategies, as 1-shot
RLVR~\citep{wang2025reinforcementlearningreasoninglarge} shows that training on a single example with replay can match the performance of training on a 1.2k-example math subset (DeepScaleR) that contains that example. However, on-policy RL methods like GRPO do not reuse samples from previous policy iterations, limiting data efficiency. A direct
response is to reuse cached trajectories and responses from earlier policies $\pi_{\theta^-}$, but this introduces
distribution shift under policy drift. RePO~\citep{li2025repo} maintains a replay buffer storing past rollouts and
retrieves off-policy samples using diverse strategies, including prioritizing samples that are high-reward or closely
aligned with the current policy. ReMix~\citep{liang2025squeeze} proposes three components to mitigate instability under
policy drift: mix-policy proximal policy gradient, a KL-convex policy constraint, and policy reincarnation. More
generally, replay can be paired with principled off-policy objectives, e.g., Trajectory Balance with
Asynchrony~\citep{bartoldson2025tba} and Soft Policy Optimization~\citep{cohen2025soft}. Note that the importance ratio
$\pi_\theta(u_t \mid s_t) / \pi_{\theta^-}(u_t \mid s_t)$ used by mix-policy updates can become extreme if policies
diverge; AR3PO~\citep{zhang2025improvingsamplingefficiencyrlvr} addresses this by recomputing token probabilities under the current $\pi_\theta$
for reused responses.
In parallel, dynamic in-context example selection selects demonstrations \emph{along} an agent trajectory~\citep{wang2025dice}, shaping the effective replay/exemplar distribution that conditions subsequent rollouts without storing full model checkpoints.

Beyond replay buffers, recent work studies how to prioritize and reuse \emph{valuable} experiences. ExGRPO~\citep{zhan2025exgrpo}
organizes and prioritizes experiences using correctness and entropy as value indicators, then employs a mixed-policy
objective to balance exploration with experience exploitation. Sample-centric progressive optimization~\citep{chen2025data}
also emphasizes per-sample prioritization via prefix-guided sampling and learning-progress weighting. For data efficiency,~\citet{sun2025difficultyreplay} combine difficulty-targeted online data selection with rollout replay, reducing
RL fine-tuning time while matching GRPO performance.

Caching can also be used not only to reuse experience across iterations, but to stabilize learning signals \emph{within}
group-based objectives. In particular, group-normalized advantages in GRPO vanish when all rollouts for a prompt yield
identical rewards, producing zero gradient. To address this, DAPO~\citep{yu2025dapo} introduces a dynamic sampling
strategy, which continues to sample new prompts and responses until the batch is fully filled with samples whose
accuracy is neither all-zero nor all-one. This over-sampling inherently incurs higher inference cost, as observed in
AR3PO~\citep{zhang2025improvingsamplingefficiencyrlvr} where they instead propose to retain correct responses from earlier
training steps in the replay buffer. When current rollouts for a sample are all incorrect, a cached correct
response will be injected from the buffer to retain the variance of rewards. Thus, incorrect rollouts receive negative advantages instead
of zero without the need for response re-generation, offering efficiency enhancement. However, open questions remain
regarding whether cached responses from $\pi_{\theta^-}$ remain valid anchors under the policy drift.

\subsection{Recomposition and Segment Reuse} \label{sec:r-recompose}

Rather than treating rollouts as atomic units, recomposition techniques decompose verified trajectories into segments or steps and recombine them into novel candidates, exploiting the modular structure of reasoning processes.
RoRecomp~\citep{li2025rorecomp} addresses RLVR's tendency toward verbosity by strategically recomposing training data into priority batches (pairing short-correct with long-incorrect responses to provide clear gradient signals for brevity) and compensation batches (replaying remaining responses from a buffer to prevent model collapse), achieving substantial reductions in reasoning length and unnecessary tool calls with minimal performance degradation. 
TreePO~\citep{li2025treepo} introduces a self-guided rollout algorithm that views sequence generation as tree-structured search with dynamic tree sampling and fixed-length segment decoding, leveraging local uncertainty to spawn branches while amortizing computation across common prefixes; segment-wise sampling alleviates KV cache burden, while tree-based segment-level advantage estimation considers both global and local signals, reducing per-update compute while preserving exploration diversity. 
Related designs reuse strong prefixes or verified anchors as reusable starting points for branching and recomposition, e.g.,
BREAD~\citep{zhang2025bread} constructs branched rollouts from expert anchors to bridge supervised traces and RL updates.
Both approaches reveal that reasoning trajectories exhibit compositional structure that can be exploited through recomposition, with the key insight being that granularity matters: response-level versus segment-level replay unlocks different efficiency gains depending on whether the goal is compressing verbose outputs or sharing redundant prefixes.

\subsection{Self-Evolving Curricula and Intrinsic Feedback} \label{sec:r-evolution}

Beyond caching and recomposing existing rollouts, self-evolution frameworks enable models to autonomously generate new training data through iterative loops where rollouts produce tasks, solutions, or agents that feed back into training.  Efforts~\citep{zhang2025path} have been made to propose a self-aware RL approach where the LLM alternates between proposing tasks and attempting to solve them. The framework introduces self-aware difficulty prediction to assess task difficulty relative to the model's current abilities and prioritize challenging yet solvable tasks, alongside self-aware limit breaking where the model recognizes capability boundaries and proactively requests minimal external data. 
Agent0~\citep{xia2025agent0} establishes symbiotic competition between two agents initialized from the same base LLM: a curriculum agent that proposes increasingly challenging frontier tasks and an executor agent that learns to solve them. The curriculum agent is rewarded by the executor's uncertainty and tool-use frequency, while the executor is trained via RL on filtered challenging problems. External tools enhance the executor's problem-solving capacity, which in turn pressures the curriculum agent to construct more complex tool-aware tasks, establishing a self-reinforcing cycle without external human-curated data. 
LANCE~\citep{wang2025language} enables LLMs to autonomously generate, clean, review, and annotate data with preference information. The model reviews seed data using constitutional principles to identify deficiencies, then generates new instruction-response pairs to address these gaps. Iterative cycles of data construction and preference-driven fine-tuning maintain high-quality generation across mathematical reasoning and general tasks. 
Complementary self-improvement loops also co-evolve reward modeling and policy learning (e.g., SPARK~\citep{liu2025spark}) or
use self-play to continuously generate new training signals in interactive agentic domains (e.g., SWE-RL~\citep{wei2025ssr}).
These approaches demonstrate that rollouts can actively shape the training distribution itself rather than serving as passive artifacts.

%% file: latex/5.discussion.tex
\section{Domains and Case Studies} \label{sec:domains}
We view a benchmark as specifying the \emph{rollout interface} on which a model policy is executed. A task instance is sampled as $x \sim \mathcal{D}$, and a rollout produces a trajectory $\tau=(x,u_{1:T},o_{1:T})$ with implicit state $s_t=(x,u_{1:t-1},o_{1:t-1})$. At each step, the model samples an output or action $u_t \sim \pi_\theta(\cdot \mid s_t)$, after which the interface returns an observation $o_t \sim P(\cdot \mid s_t,u_t)$, where $o_{1:T}=\emptyset$ in text-only settings. This induces a trajectory distribution
\begin{equation}
p_\theta(\tau \mid x) \;=\; \prod_{t=1}^{T} \pi_\theta(u_t \mid s_t)\, P(o_t \mid s_t,u_t),
\end{equation}
where $T$ is a stopping time (EOS, max length, success, or environment termination). Under this view, the benchmark determines the structure of rollouts and the form of feedback available for scoring and filtering.

We organize benchmarks into four categories based on the rollout interfaces they define:
\begin{enumerate}
\item \textbf{Verifiable language interfaces (math, code, SQL).} Text-in and text-out tasks with programmatic verification, such as normalized final-answer checks for math or execution-based checks for code and SQL.
\item \textbf{Multimodal reasoning interfaces.} Tasks with non-text inputs (vision, audio, video) where supervision relies on modality-aware verifiers, often implemented via structured answer extraction and rule-based checks.
\item \textbf{Agentic interactive interfaces.} Multi-step environments with tool or state feedback, where observations are non-empty and success is defined over trajectories and terminal environment states rather than a reference string.
\item \textbf{Agentic skill interfaces.} Environments that evaluate skill induction, library management, and cross-task transfer of reusable procedures, rather than independent episode solving alone.
\end{enumerate}

\subsection{Verifiable Language Interfaces (Math, Code, SQL)}
Math benchmarks provide a canonical verifiable interface: rollouts are typically text-only so $o_{1:T}=\emptyset$ and $\tau=(x,u_{1:T})$ reduces to a completion $y=u_{1:T}$. A deterministic final-answer verifier $V(x,y)$ yields a binary correctness signal that can be used for evaluation and as a filtering primitive in rollout-based training. A representative dataset is MATH, which contains 12{,}500 competition mathematics problems with step-by-step solutions and supports exact match evaluation after answer normalization~\citep{hendrycks2021measuring}. Standard evaluation suites for this interface include MATH500 and contest-style sets such as AIME/AMC, often alongside harder curated benchmarks such as OlympiadBench~\citep{he2024olympiadbench} (e.g., as reported in TreeRL~\citep{hou2025treerlllmreinforcementlearning}).

\paragraph{Math Case Study.}
In modern math reasoning post-training, this verifiable interface is often paired with RLVR-style objectives and exact-answer rewards, where rollout design (grouping/tree topology and sampling policy) materially affects both stability and cost. Representative systems scale verifiable training with math-specific data pipelines and RLVR-style objectives (e.g., DeepSeekMath~\citep{shao2024deepseekmath}; DeepSeek-R1~\citep{guo2025deepseek}; SEED-GRPO~\citep{chen2025seedgrposemanticentropyenhanced}), and increasingly rely on tree/group rollouts (TreeRL~\citep{hou2025treerlllmreinforcementlearning}; TreeRPO~\citep{yang2025treerpo}) and variance/uncertainty-aware curriculum or sampling (VCRL~\citep{jiang2025vcrl}) to improve reliability and efficiency.
When process supervision is available, step-wise verifiers can further densify the interface beyond terminal answer checks (e.g., Math-Shepherd~\citep{wang2023math}), shifting the rollout design emphasis toward prefix/segment-level filtering and control.

Code and SQL benchmarks define a closely related but execution-grounded interface. The model outputs $u_{1:T}$ specify a program, an edit sequence, or a query, while verification is implemented via compilation, runtime execution, and unit tests, or via database execution for Text-to-SQL. LiveCodeBench~\citep{jain2024livecodebench} is a representative benchmark suite that evaluates multiple settings including direct generation and execution-aware variants. Text-to-SQL benchmarks instantiate the same execution-grounded interface with a database engine as the verifier, enabling correctness checks based on query execution outcomes (e.g., BIRD~\citep{li2023llmservedatabaseinterface}).

\paragraph{Code/SQL Case Study.}
Execution verifiers naturally induce multi-stage rollouts (generate $\rightarrow$ compile/execute $\rightarrow$ observe failures $\rightarrow$ retry/repair), making filtering and replay particularly concrete. CodeRL~\citep{le2022coderl} and RLTF~\citep{liu2023rltf} explicitly use compilation/tests as feedback signals to shape rollouts and learning. For Text-to-SQL, recent work demonstrates that execution outcomes can be used as scalar rewards under GRPO-style updates~\citep{kulkarni2025reinforcing}, and RLVR-style systems can be built with simple rewards and strong reasoning~\citep{yao2026arctictext2sqlr1simplerewardsstrong,ali2025stateoftheartsqlreasoningmodel}. Recomposition and segment reuse are also natural in this domain, since verified patches, tests, or partial traces can be cached and recombined~\citep{li2025rorecomp}.

\subsection{Multimodal Reasoning Interfaces}
Multimodal benchmarks extend task instances beyond text, for example $x=(v,q)\sim\mathcal{D}_{\mathrm{mm}}$ where $v$ is an image or video and $q$ is a textual query. A rollout produces a trajectory $\tau=(x,u_{1:T},o_{1:T})$ and the primary object scored by the benchmark is the generated completion $y=u_{1:T}$. The defining component of the interface is the verifier, which extracts a structured answer from $y$ and compares it to annotations or rule-based checks. Recent benchmarks and data pipelines emphasize making this verification pathway deterministic and scalable, including spatially grounded video reasoning with verifiable evaluation~\citep{ouyang2025spacer}, verifiable synthesis pipelines for constructing checkable multimodal supervision~\citep{wu2025synthrl}, and multimodal post-training resources that formalize reward construction and evaluation under verifiable or semi-verifiable signals~\citep{zhang2025r1,leng2025mmr1enhancingmultimodalreasoning,li2026importance}. Additional multimodal reasoning resources follow the same interface while varying input modalities and answer formats~\citep{huang2025visionr1incentivizingreasoningcapability,shen2025vlmr1stablegeneralizabler1style, mundada-etal-2025-wildscore,surana2026musicrsbenchmarkingaudiocentricconversational}.

\subsection{Agentic Interactive Benchmarks}
Agentic benchmarks define an interactive rollout interface where the model alternates between actions and environment feedback. A task instance $x \sim \mathcal{D}_{\mathrm{ag}}$ specifies an initial context and an action space over tools or environment operations. A rollout yields $\tau=(x,u_{1:T},o_{1:T})$ with actions $u_t \sim \pi_\theta(\cdot \mid s_t)$ and observations $o_t \sim P(\cdot \mid s_t,u_t)$. Progress and correctness are mediated by tool outputs, execution results, or simulator responses, and cost is often dominated by long horizons and expensive interactions.

Software engineering agents instantiate this interface through codebase editing and test execution. Agent-RLVR provides repository and issue environments where the agent proposes patches and validates them via unit tests, producing observations such as build errors and test failures that define filtering and rewards~\citep{da2025agentrlvrtrainingsoftwareengineering}. Related evaluations commonly use SWE-Bench~\citep{jimenez2023swe} and agent-oriented wrappers such as SWE-agent~\citep{yang2024swe} and SWE-Gym~\citep{pan2024training}.

Web agents expose a tool-mediated loop over webpages and external content. BrowserGym~\citep{chezelles2024browsergym} and AgentDojo~\citep{debenedetti2024agentdojo} define tasks with actions such as clicking, typing, and navigation, and observations that reflect page state and tool outputs. ARLAS~\citep{wang2025adversarial} complements these interfaces with robustness-focused evaluation and adversarial training, including vulnerability to indirect prompt injection.

Dialogue simulators provide a text-based interactive interface with structured feedback. RLVER~\citep{wang2025rlverreinforcementlearningverifiable} uses a simulated user that updates an internal emotion score $e_t$ after each turn, inducing a verifiable trajectory-level reward from the emotion trajectory (e.g., the terminal score $e_T$) under a stopping condition defined by goal completion or a turn limit, and evaluates within the SAGE simulator framework~\citep{zhang2025sage}.

More broadly, agentic post-training increasingly couples tool-logged trajectories with multi-turn control policies and self-generated curricula, e.g., Agent0~\citep{xia2025agent0} alternates between curriculum construction and execution with tool use, producing interactive rollouts whose value depends on both terminal success and intermediate tool feedback.

Across these interfaces, the benchmark determines where feedback enters the trajectory, how verification is computed, and how rollout budgets should be allocated across depth, branching, and replay.

\input{tables/benchmark}

\subsection{Agentic Skills Benchmarks}
\label{sec:agent-skills}
Agentic skill benchmarks evaluate whether an agent can induce reusable procedures from trajectories, store them, and transfer them to new tasks. Unlike the general agentic benchmarks above, which reward \emph{capability reuse} only through independent episode solving, these interfaces reward \emph{capability reuse} through skill induction, library management, and cross-task generalization.

Web-skill suites centre on WebArena and Mind2Web (typically instantiated through BrowserGym~\citep{chezelles2024browsergym}). Agent Workflow Memory~\citep{wang2024agentworkflowmemory} abstracts sub-routines into parameterized natural-language workflows retrieved in future episodes, measuring cross-task and cross-domain transfer. Subsequent work replaces text workflows with executable programs: Agent Skill Induction~\citep{wang2025inducingprogrammaticskills} represents skills as Python functions verified via re-execution, and SkillWeaver~\citep{zheng2025skillweaver} has agents autonomously discover and refine skills into reusable APIs that transfer across models, while ReUseIt~\citep{liu2025reuseit} synthesizes reusable workflows with execution guards from both successful and failed attempts. In tool-calling environments, Agent World Model~\citep{wang2026agentworldmodel} trains agents via GRPO on ${\sim}$1{,}000 synthetic MCP-server environments and evaluates out-of-distribution on BFCLv3, $\tau^2$-bench, and MCP-Universe, and Trajectory2Task~\citep{wang2026trajectory2task} synthesizes verifiable trajectories for ambiguous, changing, and infeasible user intents through its Retail-3I benchmark. For long-horizon app workflows, SAGE~\citep{wang2025rlselfimprovingskilllibrary} integrates a programmatic skill library into GRPO-based RL with sequential rollouts on AppWorld, where the scenario-level metric directly measures within-chain skill transfer.

Memory-centric and embodied suites test whether skill representations generalize across datasets and models. MemSkill~\citep{zhang2026memskill} treats memory operations as learnable, evolvable skills via PPO-trained selection, evaluating on LoCoMo, LongMemEval, ALFWorld, and HotpotQA. SkillRL~\citep{xia2026skillrl} co-evolves a hierarchical skill bank with the policy during GRPO on ALFWorld, WebShop, and search-augmented QA tasks. Concurrent work explores complementary procedural-memory mechanisms across overlapping benchmarks: hierarchical memory with Bayesian reliability tracking~\citep{forouzandeh2025macla}, step-level and script-level procedural abstractions with cross-model transfer~\citep{fang2025memp}, dual-form reusable expertise with continuous scoring and pruning~\citep{qiu2026autorefine}, exploration-driven skill discovery with iterative feedback~\citep{yang2025exif}, trajectory distillation into reusable strategic principles~\citep{wu2025evolver}, and modular memory units for multi-agent workflow automation~\citep{han2025legomem}. Across these settings, success is measured not only by terminal task completion but also by efficiency gains from skill reuse, robustness under changing intents, and cross-task transfer of learned procedures.

\section{Failure Modes and Open Problems}
\label{sec:failures}
Rollout pipelines tend to fail in recurring, diagnosable ways, but their root causes can span multiple stages of the Generate--Filter--Control--Replay (GFCR) lifecycle. Table~\ref{tab:failures_gfcr_map} provides a troubleshooting index: each row names a symptom, suggests a \emph{primary} module to inspect first, and links to the most relevant subsections for concrete diagnostics and mitigation levers. Rows are grouped and color-coded by module for fast scanning. The \emph{Primary module} column is intentionally a starting point rather than a unique attribution---when a local fix does not resolve the symptom, follow the section pointers to audit neighboring modules and shared measurement assumptions (e.g., verifier calibration, budget accounting, and provenance).

\input{tables/failure-mode}

\paragraph{Open problems.} Several recurrent gaps cut across modules:
\begin{itemize}
\item \textbf{Verifier/judge evaluation and calibration.} We lack standardized, domain-spanning protocols for measuring verifier error rates, robustness to formatting/normalization, and how verifier calibration drifts over time~\citep{yan2025verifybench,huang2025pitfalls}. This impacts both filtering reliability (\S\ref{sec:f-verification}, \S\ref{sec:f-comparative}) and control decisions that depend on early feedback (\S\ref{sec:c-stop}).
\item \textbf{Reward misalignment and transfer.} Even with verifiable rewards, RLVR systems can learn brittle heuristics or exploit reward artifacts, raising questions about what forms of supervision best transfer across tasks and domains~\citep{shao2025spurious,gao2023scaling}. This is tightly coupled to training-signal construction (\S\ref{sec:f-supervision}) and generation diversity (\S\ref{sec:g_sampling}).
\item \textbf{Compute accounting and measurement.} Many methods report tokens but omit comparable accounting of tool calls, verifier runtime, branch/prune overhead, and replay refresh, making it difficult to assess true cost/benefit trade-offs~\citep{tu2025positionhiddencostsmeasurement}. This affects Control and Systems choices (\S\ref{sec:c-budget}, \S\ref{sec:c-systems}).
\item \textbf{Safe reuse and self-evolution.} Replay and self-evolution raise unresolved questions about provenance tracking, contamination, and how to bound the influence of self-generated data while still enabling autonomous curricula~\citep{jain2024livecodebench}. This spans Replay and Filter (\S\ref{sec:r-evolution}, \S\ref{sec:f_structural_validity}).
\end{itemize}

\paragraph{Reading guide.}
In practice, many failures arise from interactions across modules. We therefore recommend starting with the primary module indicated in
Table~\ref{tab:failures_gfcr_map}, then checking the referenced subsections for specific controls, safeguards, and measurement guidance. The visual icons appearing in the figures were created using OpenAI’s ChatGPT and are intended purely as illustrations.

%% file: tables/benchmark.tex
\begin{table}[ht]
\centering
\caption{Benchmark families viewed as rollout interfaces, with representative case studies and verifiers.}
\scriptsize
\setlength{\tabcolsep}{5pt}
\renewcommand{\arraystretch}{1.10}

\begin{tabularx}{\linewidth}{@{} l l X X @{}}
\toprule
\rowcolor{gfcrHead}
\textbf{Interface family} & \textbf{Case study} & \textbf{Interface and feedback} & \textbf{Representative benchmarks / resources} \\
\hboldline

\rowcolor{gfcrG}
\multicolumn{4}{@{}l@{}}{\textsc{\textbf{Verifiable Language}}} \\

\rowcolor{gfcrG!70}
\quad \textbf{Verifiable language} & Math &
Text-only rollout ($o_{1:T}=\emptyset$); deterministic final-answer verification $V(x,y)$ after normalization. &
MATH~\citep{hendrycks2021measuring}; OlympiadBench~\citep{he2024olympiadbench}; TreeRL/TreeRPO~\citep{hou2025treerlllmreinforcementlearning,yang2025treerpo}. \\

\rowcolor{gfcrG!55}
\quad \textbf{Verifiable language} & Code &
Program/patch/query outputs; verification via compilation, execution, and unit tests. &
LiveCodeBench~\citep{jain2024livecodebench}; CodeRL/RLTF~\citep{le2022coderl,liu2023rltf}. \\

\rowcolor{gfcrG!40}
\quad \textbf{Verifiable language} & SQL &
Query execution with a database engine; correctness by execution result. &
BIRD~\citep{li2023llmservedatabaseinterface}; execution-rewarded Text-to-SQL RL~\citep{kulkarni2025reinforcing}; Arctic-Text2SQL-R1~\citep{yao2026arctictext2sqlr1simplerewardsstrong}. \\
\hline

\rowcolor{gfcrF}
\multicolumn{4}{@{}l@{}}{\textsc{\textbf{Multimodal Reasoning}}} \\

\rowcolor{gfcrF!70}
\quad \textbf{Multimodal reasoning} & General VLM post-training &
Modality-aware verification via structured answer extraction + label/rule checks. &
Multimodal RLVR resources (R1-VL; MMR1)~\citep{zhang2025r1,leng2025mmr1enhancingmultimodalreasoning}. \\

\rowcolor{gfcrF!55}
\quad \textbf{Multimodal reasoning} & Space / spatial-video &
Rule-based or grounded evaluation to keep supervision deterministic and scalable. &
SpaceR~\citep{ouyang2025spacer};SPACEVISTA~\citep{sun2025spacevista};  InternSpatial~\citep{deng2025internspatial}; SPAR~\citep{zhang2025flatland}; VSI-Bench ~\citep{yang2025thinking}. \\
\hline

\rowcolor{gfcrC}
\multicolumn{4}{@{}l@{}}{\textsc{\textbf{Agentic Interactive}}} \\

\rowcolor{gfcrC!70}
\quad \textbf{Agentic interactive} & Software engineering &
Tool loop with non-empty observations; success measured by terminal repo/test state. &
Agent-RLVR~\citep{da2025agentrlvrtrainingsoftwareengineering}; SWE-Bench / SWE-agent / SWE-Gym~\citep{jimenez2023swe,yang2024swe,pan2024training}. \\

\rowcolor{gfcrC!55}
\quad \textbf{Agentic interactive} & Web agents &
Click/type/navigate actions; observations from page state and tool outputs. &
BrowserGym; AgentDojo; ARLAS~\citep{chezelles2024browsergym,debenedetti2024agentdojo,wang2025adversarial}. \\

\rowcolor{gfcrC!40}
\quad \textbf{Agentic interactive} & Dialogue simulators &
Turn-based interaction; reward from simulator state and terminal goal conditions. &
RLVER; SAGE~\citep{wang2025rlverreinforcementlearningverifiable,zhang2025sage}. \\
\hline

\rowcolor{gfcrR}
\multicolumn{4}{@{}l@{}}{\textsc{\textbf{Agentic Skills}}} \\

\rowcolor{gfcrR!70}
\quad \textbf{Agentic skills} & Web-skill suites &
Skill induction from trajectories; reuse via memory retrieval or programmatic re-execution. &
AWM~\citep{wang2024agentworkflowmemory}; ASI~\citep{wang2025inducingprogrammaticskills}; SkillWeaver~\citep{zheng2025skillweaver}; ReUseIt~\citep{liu2025reuseit}. \\

\rowcolor{gfcrR!55}
\quad \textbf{Agentic skills} & Tool-calling suites &
Synthetic environment generation; verifiable trajectories for non-idealized intents. &
Agent World Model~\citep{wang2026agentworldmodel}; Trajectory2Task~\citep{wang2026trajectory2task}. \\

\rowcolor{gfcrR!40}
\quad \textbf{Agentic skills} & Long-horizon \& memory-centric &
Skill-library integration with RL; procedural memory with cross-task/cross-model transfer. &
SAGE~\citep{wang2025rlselfimprovingskilllibrary}; MemSkill~\citep{zhang2026memskill}; SkillRL~\citep{xia2026skillrl}; MACLA~\citep{forouzandeh2025macla}; Mem$^\mathrm{p}$~\citep{fang2025memp}; EXIF~\citep{yang2025exif}. \\

\boldbottomline
\end{tabularx}
\end{table}

%% file: tables/failure-mode.tex
\begin{table}[!ht]
\centering
\caption{\textbf{Troubleshooting index for rollout pathologies.} Each row maps an observed failure mode to the GFCR module that is typically the best first place to intervene, and provides diagnosis/mitigation pointers.}
\vspace{2.0mm}
\label{tab:failures_gfcr_map}
\scriptsize
\setlength{\tabcolsep}{5pt}
\renewcommand{\arraystretch}{1.12}

\begin{tabularx}{\linewidth}{@{} >{\raggedright\arraybackslash}p{0.23\linewidth} >{\centering\arraybackslash}p{0.16\linewidth} Y @{}}
\toprule
\rowcolor{gfcrHead}
\textbf{Issue} & \textbf{Primary module} & \textbf{Diagnosis and mitigation pointers} \\
\hboldline

\gfcrsection{gfcrF}{Filter: scoring, verification, supervision}
\rowcolor{gfcrF!70}
Spurious signals and bias &
\GFCRF &
\textbf{Diagnosis.} Scores correlate with superficial properties (e.g., length, style, formatting); models exploit judge/verifier artifacts.\newline
\textbf{Mitigation.} Strengthen structural validity and safety gates (\S\ref{sec:f_structural_validity}); use calibrated comparative assessment with abstention where appropriate (\S\ref{sec:f-comparative}); design supervision to reduce proxy optimization (\S\ref{sec:f-supervision}). \\

\rowcolor{gfcrF!55}
Verifier brittleness and metric mismatch &
\GFCRF &
\textbf{Diagnosis.} High sensitivity to normalization and parsing; false positives/negatives under small output perturbations.\newline
\textbf{Mitigation.} Prefer deterministic, execution-based verification when available (\S\ref{sec:f-verification}); add redundant checks or consistency constraints, and incorporate process-level signals when terminal verification is noisy (\S\ref{sec:f-process}). \\

\rowcolor{gfcrF!40}
Reward hacking and over-optimization &
\GFCRF &
\textbf{Diagnosis.} Policies exploit proxy rewards (including judge idiosyncrasies) rather than improving task success; gains may not transfer and can regress under distribution shift.\newline
\textbf{Mitigation.} Prefer verifiable checks when possible (\S\ref{sec:f-verification}); improve reward/judge robustness and calibration (\S\ref{sec:f-comparative}); use supervision mappings that reduce proxy exploitation and length/style bias (\S\ref{sec:f-supervision}). \\
\hline

\gfcrsection{gfcrG}{Generate: exploration and diversity}
\rowcolor{gfcrG!65}
Coverage collapse and near-duplicate rollouts &
\GFCRG &
\textbf{Diagnosis.} Candidate sets have low diversity, high duplication, and limited disagreement, reducing effective supervision.\newline
\textbf{Mitigation.} Increase exploration and diversity controls (\S\ref{sec:g_sampling}); diversify guidance and scaffolding (\S\ref{sec:g_guidance}); allocate budget explicitly to exploration when needed (\S\ref{sec:c-budget}). \\
\hline

\gfcrsection{gfcrC}{Control: budgeting, stopping, interaction policies}
\rowcolor{gfcrC!70}
Compute efficiency and tail performance regressions &
\GFCRC &
\textbf{Diagnosis.} Early stopping and partial rollouts improve average efficiency but reduce accuracy on difficult instances.\newline
\textbf{Mitigation.} Report per-difficulty token accounting and tail metrics; use tail-aware budgeting and scheduling (\S\ref{sec:c-budget}); apply conservative continuation/stopping rules (\S\ref{sec:c-stop}); adjust rollout configuration, including length control (\S\ref{sec:c-config}). \\

\rowcolor{gfcrC!55}
Vanishing advantages in group-relative updates &
\GFCRC &
\textbf{Diagnosis.} Within-group reward variance is low, causing relative advantages to collapse and updates to weaken.\newline
\textbf{Mitigation.} Prioritize prompts with higher variance or uncertainty (\S\ref{sec:c-task}); adapt group size and sampling budget (\S\ref{sec:c-budget}); emphasize learning-value signals and group-aware weighting (\S\ref{sec:f-learningvalue}, \S\ref{sec:f-supervision}); consider replay-based variance-restoration strategies (\S\ref{sec:r-buffer}). \\

\rowcolor{gfcrC!40}
Instability in interactive search and tool use &
\GFCRC &
\textbf{Diagnosis.} Over-pruning, repeated tool failures, or looping behavior; success depends strongly on branching and retry policies.\newline
\textbf{Mitigation.} Use principled branch/prune control (\S\ref{sec:c-branch}); choose interaction topology that matches feedback latency (\S\ref{sec:g_topology}); enforce executability constraints and tool safety gates (\S\ref{sec:f_structural_validity}). \\
\hline

\gfcrsection{gfcrR}{Replay: reuse, drift, self-evolution}
\rowcolor{gfcrR!70}
Replay drift and missing provenance &
\GFCRR &
\textbf{Diagnosis.} Reused trajectories become stale as policies and environments change; recomposition can propagate unverified fragments.\newline
\textbf{Mitigation.} Apply retention policies that track recency and versioning (\S\ref{sec:r-buffer}); constrain recomposition to verified segments and re-verify upon reuse (\S\ref{sec:r-recompose}, \S\ref{sec:f-verification}); control on-policy/off-policy mixing when replay is used for training (\S\ref{sec:c-offpolicy}). \\

\rowcolor{gfcrR!50}
Unsafe self-evolution and self-amplified errors &
\GFCRR &
\textbf{Diagnosis.} Self-generated tasks or data induce distribution shift, contamination, or escalating invalid and unsafe outputs.\newline
\textbf{Mitigation.} Require strict acceptance gates and verifier-backed inclusion (\S\ref{sec:f_structural_validity}, \S\ref{sec:f-verification}); limit the share of self-generated data and enforce periodic evaluation on fixed, external benchmarks (\S\ref{sec:r-evolution}). \\

\boldbottomline
\end{tabularx}
\end{table}

%% file: latex/6_conclusion.tex
\section{Conclusion}
In this survey, we argued that rollout design is the missing link between optimization objectives and the experience that drives post-training for reasoning-oriented LLMs.
We formalized rollout pipelines with unified notation and introduced the Generate--Filter--Control--Replay (GFCR) framework to decompose them into four composable modules, and used it to synthesize recent methods across verifiable language interfaces (math, code, SQL), multimodal reasoning, interactive agentic settings, and agentic skill benchmarks that evaluate skill induction, reuse, and cross-task transfer.
Together with our criterion taxonomy (Table~\ref{tab:rollout_criteria_refstyle_noaspect}), GFCR makes rollout choices explicit and comparable, and supports a diagnostic view of recurring failure modes (Table~\ref{tab:failures_gfcr_map}) and their mitigation levers.
Looking ahead, key open problems include robust verifier/judge evaluation and calibration, principled compute accounting beyond token counts, and safe reuse and self-evolution with provenance tracking (\S\ref{sec:failures}).
We hope this perspective helps practitioners build and debug rollout pipelines, and encourages transparent reporting of rollout details that materially affect outcomes (e.g., verifier/judge configurations, budget policies, branching/pruning overheads, and replay refresh rules).